\documentclass{article}

\usepackage[english]{babel}

\usepackage{hyperref}

\usepackage[accepted]{icml2025}

\usepackage{amsmath}
\usepackage{amssymb}
\usepackage{mathtools}
\usepackage{amsthm}

\usepackage[capitalize,noabbrev]{cleveref}

\usepackage{amsthm}
\usepackage{amssymb}
\usepackage{mathtools}   
\usepackage[capitalize,noabbrev]{cleveref}
\usepackage{subcaption}
\usepackage{natbib}

\usepackage{microtype}

\theoremstyle{plain}
\newtheorem{theorem}{Theorem}[section]

\theoremstyle{definition}
\newtheorem{definition}[theorem]{Definition}

\theoremstyle{remark}

\definecolor{brinkpink}{rgb}{0.98, 0.38, 0.5}
\definecolor{blue_dark}{rgb}{0.0, 0.5, 0.69}
\definecolor{orange}{rgb}{1.0, 0.49, 0.0}

\date{\today}

\begin{document}

\twocolumn[
\icmltitle{Concept Reachability in Diffusion Models: Beyond Dataset Constraints}

\icmlsetsymbol{equal}{*}

\begin{icmlauthorlist}
\icmlauthor{Marta Aparicio Rodriguez}{icl}
\icmlauthor{Xenia Miscouridou}{icl,ucy}
\icmlauthor{Anastasia Borovykh}{icl}
\end{icmlauthorlist}

\icmlaffiliation{icl}{Department of Mathematics, Imperial College London, UK}
\icmlaffiliation{ucy}{Department of Mathematics and Statistics, University of Cyprus, Cyprus}

\icmlcorrespondingauthor{Marta Aparicio Rodriguez}{marta.aparicio-rodriguez22@imperial.ac.uk}

\icmlkeywords{Machine Learning, ICML}

\vskip 0.3in
]



\printAffiliationsAndNotice{}  % leave blank if no need to mention equal contribution

\begin{abstract}
    Despite significant advances in quality and complexity of the generations in text-to-image models, \emph{prompting} does not always lead to the desired outputs. Controlling model behaviour by directly \emph{steering}
    intermediate model activations has emerged as a viable alternative %However, we lack a clear understanding of when a prompt or alternative method will successfully produce a desired output. 
    allowing to \emph{reach} concepts in latent space that may otherwise remain inaccessible by prompt. 
    %We hypothesise that the training dataset plays a key role in the success or failure of generating or \emph{reaching} a concept, and 
    In this work, we introduce a set of experiments to deepen our understanding of concept reachability. We design a training data setup with three key obstacles: scarcity of concepts, underspecification of concepts in the captions, and data biases with tied concepts. 
    %We show that while prompting produces successful results in balanced datasets, steering methods become more effective at reaching the desired concepts as each of these challenges becomes more pronounced in the dataset. %We also observe trends suggesting a phase transition in concept reachability as the presence of a concept increases across the dataset, and significant limitations when concepts are underspecified, highlighting the importance of carefully curating train datasets and selecting the appropriate generation methods in order to ensure successful concept reachability.
    Our results show: (i) concept reachability in latent space exhibits a distinct phase transition, with only a small number of samples being sufficient to enable reachability, (ii) \emph{where} in the latent space the intervention is performed critically impacts reachability, showing that certain concepts are reachable only at certain stages of transformation, and (iii) while prompting ability rapidly diminishes with a decrease in quality of the dataset, concepts often remain reliably reachable through steering. Model providers can leverage this to bypass costly retraining and dataset curation and instead innovate with user-facing control mechanisms.
\end{abstract}

\section{Introduction}

\begin{figure}[htb]
    \centering
    \includegraphics[width=0.95\columnwidth]{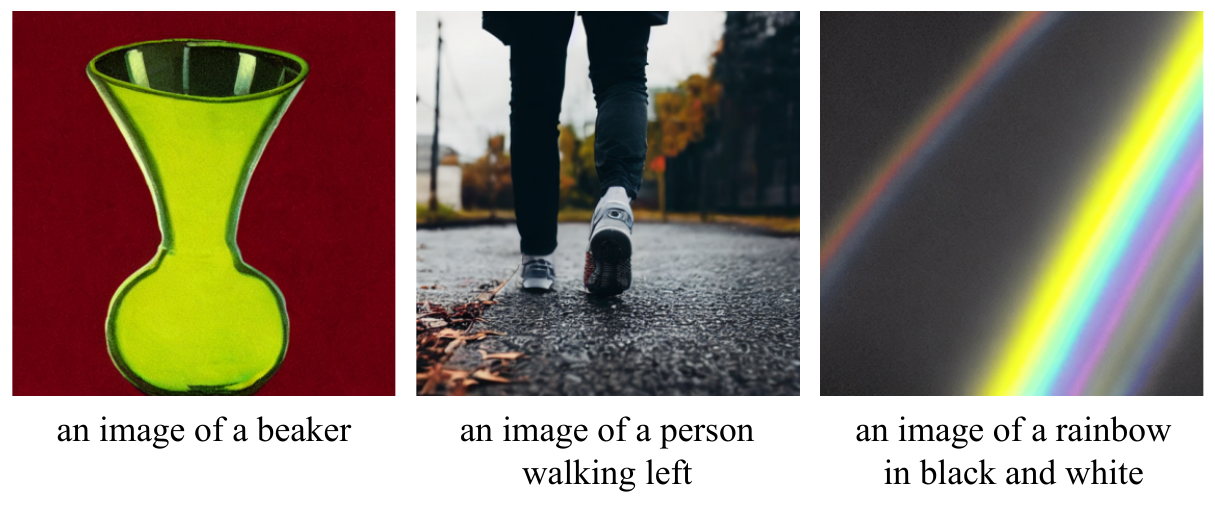}
    \caption{Images generated by Stable Diffusion \citep{latent_diff} that fail to produce the desired outcome due to hypothesised dataset limits: (L) a scarce concept, (C) underspecification in a caption, (R) biases. See Appendix \ref{app:train_stable} for details. %We identify a low number of images containing beakers in the train set relative to the total dataset size (Appendix \ref{app:train_stable}). Moreover, we find it is difficult to control the direction of movement of a person walking, which we hypothesise is due to a lack of labelling this property in captions. Additionally, the model struggles to generate black-and-white images of rainbows, despite handling black-and-white colour representation in other contexts, indicating a dataset bias.
    } \label{fig:stable_diff}
\end{figure}

The scaling of diffusion models \citep{og_diff, ncsn, diffusion, score} has significantly expanded their capacity to store and generate vast amounts of complex concepts. While prompts have become the de-facto manner to control the model output, there are numerous examples when simply prompting falls short (see Figure \ref{fig:stable_diff}). In addition to learning the visual and spatial components of concepts in images, text-to-image models must correctly associate the concept in an image with its corresponding semantic description in the caption \citep{compbench, ghosh, mouse_chase}. When this alignment fails, even overspecifying and re-prompting may fail to generate the target image.  
%As a result, the model's ability to generate an image containing a certain concept may not necessarily align with its ability to generate that concept through prompting.

Prior work has shown that, as an alternative to prompt-based sampling, one can operate directly on a model's representation level \citep{semantic, task_vectors,  concept_algebra, reft}. In particular, by editing specific activations, the sampling trajectory of diffusion models can be adjusted towards a particular target \citep{steer_seed, cross_attention, h_space_safe}. %Similarly, in generative models, recent work on interpretability has shown that models have monosemantic features which can be activated to control outputs \todo{(SAE stuff, Golden Gate Claude)}. 
%However, existing work has shown that certain features are more easily activated and controlled \citep{steerability}. %Additionally, the ability to generate combinations of features is largely unexplored in diffusion models. 
% So if the model does not understand the prompt vs if it understands the prompt but does not know what the concept is? Could look at train set frequency or accuracy of concepts that make up the prompt separately?
While these works show that it \emph{is} possible to steer towards certain output concepts, we lack a concrete framework to understand the complexity of guidance, or in other words the \textbf{reachability of concepts}. When can a concept be reliably accessed through prompting? If prompting fails, under what conditions can steering reliably reach the concept? And ultimately, what factors render a concept entirely unreachable despite it being in some way present in the training data? %When prompting fails, when does generating a concept become feasible through steering? When is it not possible to reach that concept via any means? 
%We address these questions by exploring how different flaws in the train set impact reachability of concepts.

Existing work has shown that the dataset structure plays a key role in reachability. For GANs and VAEs,  \citet{dots} conclude that biases in the train set will influence the generation ability of models, regardless of the architecture or training algorithm implemented. In diffusion models, \citet{mouse_chase} show that certain characteristics such as balanced datasets help mitigate the failure of a generated image to match the target prompt. However, the complexity of the real world complicates the construction of a balanced training dataset that fully captures the true data-generating process. 
Figure \ref{fig:stable_diff} highlights scenarios in which prompting fails to produce the correct output. %, despite it being reachable by steering.
%We design a synthetic experimental setup that is both simple and highly controllable, enabling us to understand in detail the mechanics of concept reachability. We centre our work around the following three dataset limitations:
Underlying these scenarios are three core dataset limitations: 
\begin{enumerate}
    \item Scarcity of concepts: real-world datasets often exhibit an uneven distribution of concepts, with some being underrepresented or appearing infrequently. Such scarcity can hinder a model’s ability to effectively learn these concepts.
    \item Underspecification of captions: real-world datasets often include captions that do not describe \emph{all} the concepts present in an image. This lack of detailed annotation can constrain a model’s ability to accurately associate unmentioned concepts with their corresponding semantic representations. % generate concepts not mentioned. %We vary the level of specification in captions in the train set and evaluate how often we can guide the generation towards unspecified concepts 
    \item Biases: real-world datasets frequently exhibit biases, where certain concepts consistently \emph{co-occur}. This inherent correlation can make it challenging for models to disentangle these concepts, limiting their ability to independently generate or represent them. %We vary the strength of a bias by increasing the images of one concept, evaluating how often it is possible to break the bias 
\end{enumerate}
Real-world images are often highly complex, with many intertwined factors that can make it challenging to isolate the exact conditions under which a model successfully reaches a concept. To address this, we work within a synthetic framework that allows us to systematically vary the structure of the dataset (for details, see Figure \ref{fig:blocks} and Section \ref{sec:problems}). 
This enables us to understand in detail how the mechanics of concept reachability are influenced by the above-mentioned three scenarios. To verify the generality of our main conclusions, we analyse the impact of the same scenarios on real-world data, including Stable Diffusion \citep{latent_diff} and CelebA \citep{celeba}. We list our contributions as:
\begin{itemize}
    %\item Improvement in reachability by steering: we compare different methods and notice an increase in reachability by steering in scenarios where prompting is inconsistent due to scarcity, underspecification and biases in the dataset. In setups when prompting is effective, steering produces similar reachability values.
    \item We show that even minor dataset limitations severely decrease the effectiveness of prompts, highlighting an inherent weakness in relying solely on this method of model control.
    \item We demonstrate that concepts remain accessible within the latent space, even under highly corrupted dataset conditions. 
    \item We demonstrate a phase shift in reachability: a rapid increase in reachability can be observed as the number of images containing a concept is increased beyond a concept-agnostic, low threshold.
    %\item 
    \item We identify when concepts cannot be reached: when a concept is not specified in the captions of the train set, models are unable to disentangle, and hence reach, the concepts effectively. 
\end{itemize}

Our work demonstrates both the limits of prompting and the \emph{resilience} of latent space interventions under three commonly present dataset limitations. Our work suggests that instead of curating new datasets or retraining models from scratch, model providers can enable users to reach concepts through novel control mechanisms. By shifting focus from data curation to user-driven model steering, providers can enhance model usability, robustness, and accessibility in ways that go beyond the limitations of prompting.\footnote{Code is available at \url{https://github.com/martaaparod/concept_reachability}.}

%Overall, our work aims to shed light on how limitations in a dataset can lead to varying levels of reachability. We present valuable insight into when alternative methods to prompting will successfully reach concepts. We generally observe that when a prompt fails, steering is a powerful alternative. Thus, when users struggle to generate a concept, model providers may not need to retrain or fine-tune a model, but instead offer users access to other methods in order to achieve reachability.

\section{Prior Work}
\paragraph{Control in LLMs}
%Having seen trillions of tokens during training, large language models can be seen as information compressors, raising the question of how do we extract target information. 
Large language models (LLMs), having been trained on vast amounts of data, can be conceptualised as powerful information compressors, raising the challenge of how to effectively extract task-specific information. Various methods have been proposed to address this challenge without requiring fine-tuning of all model parameters. Some approaches focus on fine-tuning a smaller subset of weights \citep{lora, peft, task_vectors, reft}, while others enhance performance by optimising the input for specific tasks \citep{ptuning}. Existing work additionally explores modifying a model's activations by introducing steering vectors at specific layers \citep{bau_paris, bau_function_vectors, contrastive_add, linear_structure_llms, steering_act_eng, inference_time_intervention}. In particular, recent advancements in Sparse Autoencoders demonstrate their potential in capturing interpretable latent representations, enabling fine-grained control and enhanced interpretability in language models \citep{saes_interpretable, goldengateclaude}.

\paragraph{Steering text-to-image models}
Similar to LLMs, previous works have successfully steered diffusion models to generate rare concepts, compose concepts and manipulate specific attributes in an image. 
\citet{steer_seed} optimise the initial random seed to address generation failures and produce desired outputs. Other methods focus on modifying the U-net output \citep{concept_algebra, bau_lora} or adjusting activations of particular layers. Research has shown that editing the generation process can be achieved by adding vectors or fine-tuning the cross-attention layers that inject semantic information into the model \citep{prompt_to_prompt, cross_attention, kldivergence_prompt, editing_text_proj}. Furthermore, semantically meaningful directions in the bottleneck layer of the U-net have been identified as a means to steer the generation effectively \citep{semantic, h_space_safe, h_pca}.

\paragraph{Data attribution methods} The study of how training data influences a model’s output \citep{kohliang, tracin, trak, influence_survey}, particularly in real data, is closely intertwined with the broader question of how dataset constraints govern reachability. Data attribution techniques have been extended to image generation in diffusion models by analysing the sampling dynamics \citep{journey, trak2}, or by examining intermediate checkpoints \citep{tracin_diffusion} or hidden representations obtained during the training process \citep{montrage}. While these approaches provide valuable insights, a significant practical challenge persists: rigorously evaluating the accuracy of influence estimation methods. To address this, \citet{evaluating_influence} introduces a methodology for constructing datasets explicitly influenced by known datapoints, thereby establishing a framework for empirical evaluation. In our work, we adopt a synthetic dataset, where the influence relationships are known by design.

\paragraph{Analysis of reachability in generative models}
Underlying the question of whether we can reach a certain concept lies the ability of the model to properly compose concepts seen during training. Research assessing this compositionality through prompting concludes that models can compose latent factors in novel ways if trained on sufficiently diverse data or for extended periods \citep{interpolation, composition}. Studies have also explored the learning dynamics that shape a model's generalisation ability. \citet{rule_extrapolation} observe a simplicity bias in LLMs, where the learning process prioritises simpler tasks earlier in training. Additionally, \citet{composition2} identify sudden transitions where diffusion models rapidly acquire the ability to generate specific concepts. In practice, failure cases do remain, with the work of \citet{mouse_chase} analysing the impact of the underlying data distribution and lack of coverage of unique phenomena.  Furthermore, steering vectors in LLMs have been found to sometimes produce unreliable or even counterproductive results \citep{steerability}. Building on these insights, our work investigates the factors contributing to the unreliability of reachability methods.

\section{Background}

\subsection{Diffusion Models}

Denoising diffusion probabilistic models \citep{diffusion} approximate the distribution $p_{data}(\mathbf{x})$ that gives rise to a collection of data points $\mathcal{X}$. During training, noise is added to images $\mathbf{x}_0$ from a train set $\mathcal{X}$ to give latents $\mathbf{x}_t = \sqrt{\bar{\alpha}_t} \mathbf{x}_{0} + \sqrt{1 - \bar{\alpha}_t} \boldsymbol{\epsilon}_t$, for $t \in [0, T]$ and appropriate constants $\bar{\alpha}_t$ dependent on a noise schedule. A U-net \citep{unet} $\boldsymbol{\epsilon}_\theta$ is trained to match the added noise by minimising the loss function
\[
\mathcal{L} = \mathbb{E}_{t \sim [1, T], \mathbf{x}_0, \boldsymbol{\epsilon}_t} \| \boldsymbol{\epsilon}_t - \boldsymbol{\epsilon}_\theta (\mathbf{x}_t, t) \|^2.
\]

We implement text-to-image diffusion models $\boldsymbol{\epsilon}_\theta (\mathbf{x}_t, t, y)$, that additionally condition generation on a text prompt $y \in \mathcal{Y}$, passed through a text encoder $\mathcal{E}$ and inputted into the U-net through cross-attention layers \citep{attentionisallyouneed, latent_diff, dalle2, glide, cross_attention}. The train set is comprised of image-caption pairs $(\mathbf{x}, y) \in \mathcal{X} \times \mathcal{Y}$. The sampling process involves the denoising of a latent $\mathbf{x}_T \sim \mathcal{N} (\mathbf{0}, \mathbf{I})$ conditioned on an input prompt $y^\prime$. See Appendix \ref{app:hyperparameters} for details on our choice of architecture and hyperparameters.

\subsection{Concepts} \label{sec:concepts}
%Structure datasets in terms of factors
We introduce assumptions on the underlying structure of our data, following a similar approach to \citet{dots, disentanglement, composition, composition2, concept_algebra}. We assume that the images in the dataset are generated by a set of factors, such as object identity, colour, position or texture. Each factor can take certain values, which we denote as \emph{concepts}.
%Define factors
Formally, we say there exists a set of $n$ concept variables $\mathbf{F} = \{F_1, F_2, \dotsc, F_n\}$, that define the image $\mathbf{x} \sim p_{data}(\mathbf{x})$. Each of the variables $F_i$ are sampled from their respective distributions $p(F_i)$. %, and their joint distribution is given by $p(\mathbf{F}) = p(F_1) p(F_2 | F_1) \dotsc p(F_n | F_1, \dotsc, F_{n-1})$. 
We denote the set of possible values the variables $F_i$ can take as $\mathcal{F}_i$, which can be discrete or continuous. The \emph{space of concepts} is then defined as the set $\mathcal{F} = \mathcal{F}_1 \times \mathcal{F}_2 \times \dotsm \times \mathcal{F}_n$ containing all possible combinations the factors can take.

Each combination of values $(f_1, f_2, \dotsc, f_n) \in \mathcal{F}$ uniquely determines an image $\mathbf{x}$, where this relation is defined by an injective function $g : \mathcal{F} \to \mathbb{R}^{{W}\times{H}\times C}$ that transforms the tuple $(f_1, f_2, \dotsc, f_n)$ into an image $\mathbf{x}$ with the target concepts $f_1, f_2, \dotsc, f_n$. This function determines the distribution of $\mathbf{x}$ when conditioned on a combination of factors.

Throughout our work, we focus on a subset of factors assumed to be identifiable for all images sampled from $p_{data}(\mathbf{x})$. Without loss of generality, we assume these are the first $m < n$ factors $\mathcal{F}_1, \mathcal{F}_2, \dotsc, \mathcal{F}_m$. We refer to the corresponding values of these factors, $(f_1, f_2, \dotsc, f_m)$, as \emph{concepts of interest}. We use the notation $[f_{i_1}, f_{i_2}, \dotsc, f_{i_l}]_\mathcal{X}$, where $i_k \in \{1, 2, \dotsc, m\}$ for $k = 1, 2, \dotsc, l$, to refer to subsets of the dataset that share the concepts $f_{i_1}, f_{i_2}, \dotsc, f_{i_l}$ for the factors $\mathcal{F}_{i_1}, \mathcal{F}_{i_2}, \dotsc, \mathcal{F}_{i_l}$, and have no restrictions along other factors.

We further assume that captions capture the expressiveness of the concepts of interest in an image. For an image $\mathbf{x}$ containing concepts of interest $(f_1, f_2, \dotsc, f_m)$, we assume there is an injective function $h: \mathcal{F}_1 \times \mathcal{F}_2 \times \dotsm \times \mathcal{F}_m \to \mathbb{R}^l$ such that $h(f_1, f_2, \dotsc, f_m)$ is a string containing the semantic information relevant to the concepts $f_1, f_2, \dotsc, f_m$. Therefore, image-caption pairs $(\mathbf{x}, y) \in \mathcal{X} \times \mathcal{Y}$ satisfy that the concepts of interest of $\mathbf{x}$ and the tuple $h^{-1} (y)$ are equal.

\subsection{Problem Setting} \label{sec:problems}
%Generating images via prompting has been shown in certain scenarios to fail to produce the intended description \citep{mouse_chase}. Research has shown that in some of these cases, with proper guidance during the generation process, the model can successfully produce the desired outcome \citep{steer_seed}. This suggests that errors in image generation can stem from the model's inability to accurately map the semantic information in prompts to the correct visual representations. However, we still lack an understanding of when a model requires further guidance to produce the desired output and when it is unable to do so via any alternative method to prompting.

\begin{figure}[htb]
    \centering
    \includegraphics[width=1\columnwidth]{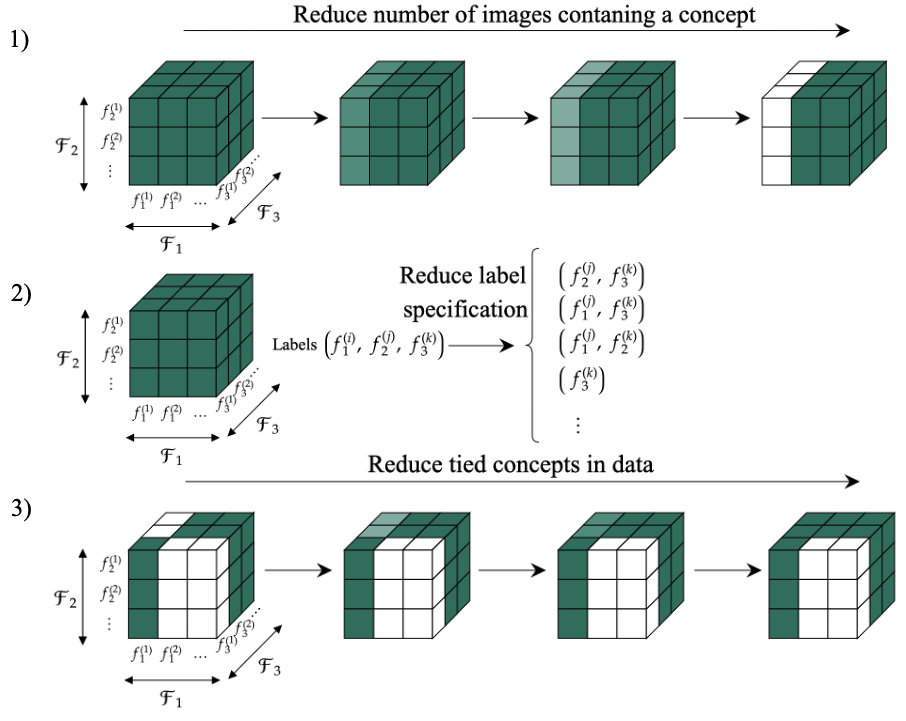}
    \caption{Visualisation of the structure of the dataset according to concepts of interest (in the diagram, three). Each block in the cube represents the collections of images in $[f_1^{(i)}, f_2^{(j)}, f_3^{(k)}]_\mathcal{X}$. A darker shade represents a higher number of images in the block. Data modifications are as described in Section \ref{sec:problems}. 
    % 1) Gradually decrease the number of images containing a concept. 2) Reduce the level of detail in the captions of train data. 3) Starting from a biased dataset where two factors are always together in the train set (in the diagram, $f_1^{(1)}$ and $f_3^{(1)}$), gradually increase the images containing only one of them.
    }
    \label{fig:blocks}
\end{figure}

Our study examines how the presence and relationships within training data influence the ability of diffusion models to reach concepts (Figure \ref{fig:blocks}). To achieve this, we systematically vary the dataset used to train a model, and observe the evolution of reachability for specific concept combinations. We vary the following: 
\begin{enumerate}
    \item Scarcity of concepts: starting with a balanced dataset, the presence of individual concepts is progressively reduced. We evaluate how the reachability of combinations containing the underrepresented concept is affected. %Underrepresentation of certain concepts in datasets is a common issue, and we examine how varying the frequency of a concept within the training data impacts the reachability of combinations involving that concept.
    %The presence of a concept across a dataset is a measurable property in datasets containing real images. However, due to the complexity of the data, it is often infeasible to identify all the concepts across the entire dataset and systematically reduce their presence to study the impact on generation.
    \item Underspecification of captions: beginning with a fully annotated dataset, semantic information relevant to certain factors is removed. The impact of this specification reduction on the model's ability to reach complete concept combinations is assessed. %Concepts in an image may not always be explicitly referred to in the corresponding caption in the training set. We vary the level of detail seen during training and analyse how it impacts reachability. %Our experiments provide insights into how training datasets can be labeled differently to improve generation, and to what extent adding more detail to a prompt improves the ability to generate the intended target image. %Sometimes adding more details to the prompt will not work
    \item Biases: from a dataset where two concepts are consistently paired, we incrementally introduce data containing only one of the concepts. We analyse how the model’s ability to independently reach each concept evolves. %Biases in real-world datasets often result in certain concepts being consistently paired together. We analyse how effectively we can disentangle such concepts, providing insights into addressing biases in datasets.
\end{enumerate}

%\begin{figure}[htbp]
%    \centering
%    \includegraphics[width=1\columnwidth]{main_plots/experiments_blocks.png}
%    \caption{Visualisation of the data modifications implemented. In all cases the complete dataset is assumed to have factors uniformly distributed. A) Gradually decrease the number of images containing a concept. B) Reduce the level of detail in the captions of train data. C) Starting from a biased dataset where two factors are always together in the train set (in the diagram, $f_1^{(1)}$ and $f_3^{(1)}$), gradually increase the images containing only one of them.}
%    \label{fig:experiments_blocks}
%\end{figure}

\subsection{Reachability}

%Given a trained diffusion model, in order to generate an image with concepts $(f_1, f_2, \dotsc, f_m)$, one will typically use the prompt $y$ satisfying $h^{-1}(y) = (f_1, f_2, \dotsc, f_m)$ to condition the generation process.

In this section, we introduce the notion of reachability and key definitions used in subsequent sections.

%Throughout our work, we use alternative methods to achieve the target image that do not require the use of a prompt \citep{semantic, h_space_safe, kldivergence_prompt}.

\begin{definition}[Concept Reachability]
    Given a target combination of concepts $(f_1, f_2, \dotsc, f_m)$ and method $M$ to access these, we define the \emph{reachability} of $(f_1, f_2, \dotsc, f_m)$ as the \emph{accuracy} or proportion of images produced by $M$ that contain the concept combination $(f_1, f_2, \dotsc, f_m)$. %We denote this as \[\mathcal{R}_M (f_1, f_2, \dotsc, f_m).\]
\end{definition}

%\begin{definition}
%    The reachability of a concept $f_i$ of the factor $\mathcal{F}_i$ is defined as 
%    \[
%    \max_M \mathbb{E}_{f_j \in \mathcal{F}_j : j \neq i} \mathcal{R}_M ((f_1, \dotsc, f_i, \dotsc, f_m)).
%    \]
%\end{definition}

%Additionally, we consider how often a concept or combination of concepts is seen across the dataset.
%\begin{definition}
%For a combination of concepts $(f_{i_1}, f_{i_2}, \dotsc f_{i_j})$, $\{i_1, i_2, \dotsc, i_j \} \subset \{1, \dotsc, m \}$, we refer to the \emph{presence} of $(f_{i_1}, f_{i_2}, \dotsc , f_{i_j})$ across $\mathcal{X}$ as the proportion of images in the dataset containing these concepts. That is,
%\[
%p_\mathcal{X} (f_{i_1}, f_{i_2}, \dotsc , f_{i_j}) = \frac{| [f_{i_1}, f_{i_2}, \dotsc, f_{i_j}]_\mathcal{X} |}{| \mathcal{X} |},
%\]
%where $| \cdot |$ denotes the number of elements in a set.
%\end{definition}

%ID vs OOD
We also identify cases during generation in which outputs are out-of-distribution.
\begin{definition}[Out-of-distribution]
    Given a model trained on a dataset $\mathcal{X} \times \mathcal{Y}$ with concept function $g$ as defined in Section \ref{sec:concepts}, a generated image $\mathbf{x}^*$ with concepts of interest $(f^*_1, f^*_2, \dotsc , f^*_m)$ is out-of-distribution (OOD) if $[f^*_1, f^*_2, \dotsc , f^*_m]_\mathcal{X} = \emptyset$.
\end{definition}

%\begin{definition}
%    Given a set of concepts of interest $f_1, f_2, \dotsc f_m \in \mathcal{F}_1 \times \mathcal{F}_2 \times \dotsc \times \mathcal{F}_m$, the presence of $f_1, f_2, \dotsc f_m$ across the dataset $\mathcal{X}$ is defined as 
%    \[
%    p_\mathcal{X} (f_1, f_2, \dotsc f_m) = \frac{| [f_1, f_2, \dotsc f_m]_\mathcal{X} |}%{| \mathcal{X} |}
%    {\max_{f^\prime_i \in \mathcal{F}_i, i = {1, 2, \dotsc, m}} | [f^\prime_1, f^\prime_2, \dotsc f^\prime_m]_\mathcal{X} |},
%    \]
%    where $| \cdot |$ denotes the number of elements in a set.
%\end{definition}

We additionally distinguish between two mechanisms that result in unseen combinations. This distinction provides a valuable framework for understanding differences in reachability levels in OOD generalisation scenarios, as in Sections \ref{sec:number} and \ref{sec:tied_concepts}. In particular, we consider a model’s ability to combine known factors into a new configuration and model's ability to transfer knowledge of attributes from one positional context to another.

\begin{definition}[Compositionally out-of-distribution]
    Given an out-of-distribution combination of concepts $F_o = (f_{1}, f_{2}, \dotsc, f_{m}) \in \mathcal{F}_{1} \times \mathcal{F}_{2} \times \dotsm \times \mathcal{F}_{m}$, we say it is \emph{compositionally out-of-distribution} if for every concept $f_{j}$ in the combination, there exists a concept combination in the train set whose $j$th component is $f_{j}$.
\end{definition}

%This refers to a model’s ability to combine known factors into a new configuration.

\begin{definition}[Positionally out-of-distribution]
    Given an out-of-distribution combination of concepts $F_o = (f_{1}, f_{2}, \dotsc, f_{m}) \in \mathcal{F}_{1} \times \mathcal{F}_{2} \times \dotsm \times \mathcal{F}_{m}$, we say it is \emph{positionally out-of-distribution} if it is not compositionally out-of-distribution, and there exists a permutation $\rho : \{ 1, 2, \dotsc , m\} \to \{ 1, 2, \dotsc , m\}$ such that $F^\prime_o = (f_{\rho(1)}, f_{\rho(2)}, \dotsc, f_{\rho(m)})$ is seen during training.
\end{definition}

%This refers to target combinations that become seen under a reordering of their components, reflecting the model's ability to transfer knowledge of attributes from one positional context to another. %(for example, for our dataset defined in Section \ref{sec:dataset}, generalising colour or shape from front to back position).

\section{Methodology}

\subsection{Dataset} \label{sec:dataset}
%\todo{We design a synthetic experimental setup that pursues simplicity and controllability while preserving the essence of the phenomenon of interest, i.e., compositional generalization. Specifically, our data-generating process tries to mimic the structure of the training data used in text-conditioned diffusion models by developing pairs of images representing geometric objects and tuples that denote which concepts are involved in the formation of a given image; say that we have a similar setup to XXX papers, but change XXX}

We design a synthetic experimental setup that pursues controllability over the tasks outlined in Section \ref{sec:problems}. %Specifically, our data-generating process mimics the structure of training data used in text-conditioned diffusion models by developing pairs of images representing geometric objects and captions that denote which concepts are of interest in the formation of a given image.
Our setup is similar to the work of \citet{dots, disentanglement, dnns_shortcut, first_principles, composition, interpolation, composition2, mouse_chase}. %This relation allows us to analyse more complex interactions in images.
We use a dataset that contains images of coloured shapes on a black background, where one shape is partially covered by the other (Figure \ref{fig:trainset}). %Each image is determined by back shape, back colour, location of back shape, front shape, front colour and location of front shape.  Our concepts of interest are chosen to be back colour, back shape, front colour and front shape, and allow for the position of the shapes to vary uniformly.
See Appendix \ref{app:data} for further details.

\begin{figure}[htbp]
    \centering
    \includegraphics[width=1\columnwidth]{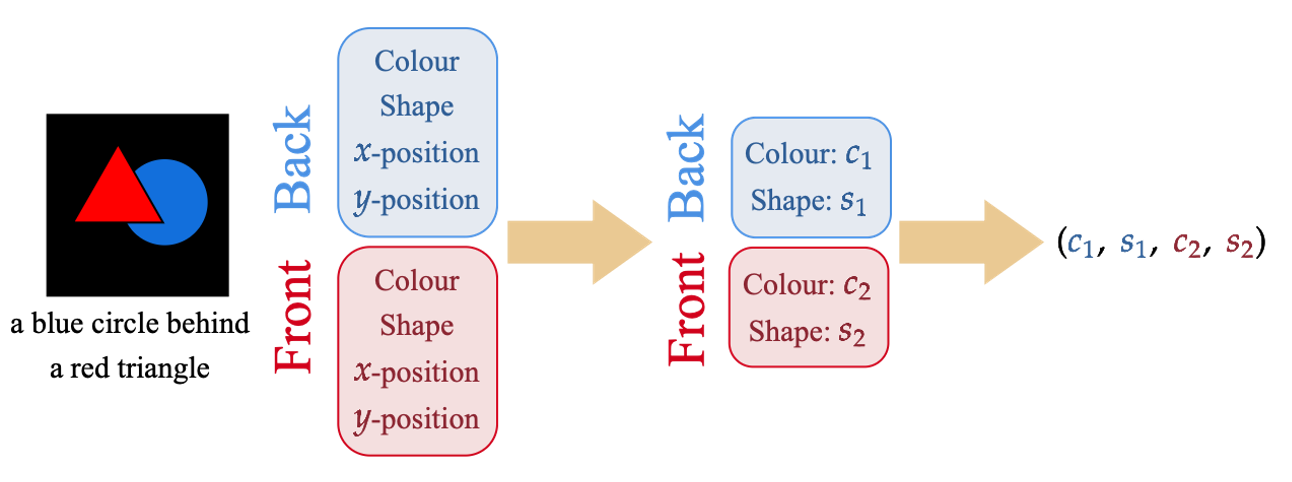}
    \caption{Concepts of interest in the dataset. The concepts of an image are summarised as the tuple $(c_1, s_1, c_2, s_2)$, where each of these positions refers to back colour, back shape, front colour and front shape respectively. In the diagram, this is $(blue, circle, red, triangle)$.}
    \label{fig:concepts_trainset}
\end{figure}

Let $\mathcal{S} =$ \{circle, triangle, square\} and $\mathcal{C} =$ \{red, green blue\}. The images in our dataset are labelled using the captions, ``a \{$c_1$\} \{$s_1$\} behind a \{$c_2$\} \{$s_2$\}", where $s_1, s_2 \in \mathcal{S}$, and 
$
(c_1, c_2) \in (\mathcal{C} \times \mathcal{C}) \setminus \{(c, c) : c \in \mathcal{C}\}.
$
That is, all combinations of any two shapes and any two colours in $\mathcal{S}$ and $\mathcal{C}$ are admitted, excluding images containing two shapes of the same colour. We refer to the concepts of interest of images as $(c_1, s_1, c_2, s_2)$ (Figure \ref{fig:concepts_trainset}). We create the dataset such that each of the combinations of concepts $(c_1, s_1, c_2, s_2)$ is originally seen an equal number of times. %This allows us to control the challenges outlined in Section \ref{sec:problems} and isolate the impact that they have in the distribution approximated by the diffusion model, as shown in Section \ref{sec:reachability}.

Our original dataset is comprised of 54 combinations of shapes and colours $(c_1, s_1, c_2, s_2)$, each containing 1000 images. During experiments where the number of images containing certain concept combinations is varied, we preserve the total size of the dataset by adjusting the size of the remaining combinations.

\subsection{Steering} \label{sec:steering}

We employ implementations that add a constant vector to a specific layer of the U-net during sampling. These methods consist of two stages: the optimisation of a concept vector using images that contain the target concept combinations, and the addition of the optimised concept vector during sampling. Motivating our choices from existing work on steering \citep{semantic, kldivergence_prompt, h_space_safe}, we consider two spaces in which to implement steering: the text encoding of the prompt and the bottleneck layer of the U-net.

Let $\boldsymbol{\epsilon}_\theta$ denote a conditional U-net trained on the dataset $\mathcal{X}\times \mathcal{Y}$. For a given combination of concepts $(f_1, f_2, \dotsc, f_m)$, the model can generate an image containing this combination by using the prompt $y_e = h(f_1, f_2, \dotsc, f_m)$. Alternatively, we sample from a starting prompt $y_s$ and steer the generation process towards the desired combination of concepts $(f_1, f_2, \dotsc, f_m)$. Note that $y_s$ may not satisfy $y_s = y_e$, or that the image produced by the model may not accurately represent the target concepts of interest. Thus, we use steering to enable the model to reach the desired output more accurately.
Let $\mathcal{Z}$ be a collection of images containing the concept combination of interest. Using the starting prompt $y_s$, we create the image-label pairs $(\mathbf{x}_0, y_s)$ for every $\mathbf{x}_0 \in \mathcal{Z}$.

\paragraph{Prompt space} Let $\mathcal{E}(y_s)$ be the text embedding outputted by the text encoder for the prompt $y_s$. We modify the sampling trajectory of $y_s$ by replacing $\mathcal{E}(y_s)$ by $\mathcal{E}(y_s) + \mathbf{v}_p$, where $\mathbf{v}_p$ is a vector of the same dimensionality as $\mathcal{E}(y_s)$. We refer to the output of the U-net after this modification as $\boldsymbol{\epsilon}_\theta (\mathbf{x}_t, t, y, \mathbf{v}_p)$ (Figure \ref{fig:steering_unet}A). To obtain the vector $\mathbf{v}_p$, the following loss is minimised:
\[
L_p = \mathbb{E}_{t \sim [1, T], (\mathbf{x}_0, y_s), \boldsymbol{\epsilon}_t} \| \boldsymbol{\epsilon}_t - \boldsymbol{\epsilon}_\theta (\mathbf{x}_t, t, y_s, \mathbf{v}_p) \|^2.
\]
Note that the weights of $\boldsymbol{\epsilon}_\theta$ are frozen. The vector $\mathbf{v}_p$ therefore accounts for the mismatch between the semantic information contained in $y_s$ and the visual information in the noisy latents obtained from images in $\mathcal{Z}$.

\paragraph{h-space} We also consider steering of the starting prompt $y_s$ on the $h$-space, following the implementation of \citet{h_space_safe}. Let $\boldsymbol{\epsilon}_\theta (\mathbf{x}_t, t, y_s, \mathbf{v}_h)$ denote the output of the U-net when inputted the noisy image $\mathbf{x}_t$, the prompt $y_s$ and the vector $\mathbf{v}_h$, which is added to the bottleneck layer output (Figure \ref{fig:steering_unet}B). The vector $\mathbf{v}_h$ is optimised to minimise the loss:
\[
L_h = \mathbb{E}_{t \sim [1, T], (\mathbf{x}_0, y_s), \boldsymbol{\epsilon}_t} \| \boldsymbol{\epsilon}_t - \boldsymbol{\epsilon}_\theta (\mathbf{x}_t, t, y_s, \mathbf{v}_h) \|^2.
\]

Similarly to steering on the prompt space, the vector $\mathbf{v}_h$ is optimised to reconcile the mismatch between the semantic information in $y_s$ and the visual concepts in the images.

%In particular, for every direction $\mathbf{v}_h$ found, we choose to initialise our optimisation at $\mathbf{v}_h = \mathbf{0}$, and use a collection of 140 images as $\mathcal{X}$. Additionally, in our setup the bottleneck of the U-net satisfies $\mathbf{v}_h \in \mathbb{R}^{128 \times 8 \times 8}$.

%\todo{Describe the analysis we want to do in several concise bullet points.}

%What we want to answer:
%Baseline: Can our method steer the model towards a concept?
%Which concepts are harder to steer to than others? Does this correlate with biases in our dataset?
%Can we apply this to more complex datasets?
%(maybe) How does this compare with other methods?

\subsection{Evaluation Method}
%\todo{How do we eval the outputs. Something a la: However, the end goal of steering is not to obtain a small loss during steering; it is to find images that are representing the concept accurately; hence we need more metrics than just the steering optimisation loss.}

Reachability, as defined in Section \ref{sec:reachability}, necessitates the identification of the concepts of interest within sampled images. Identifying concepts in images on a large scale can prove a challenging task, with most real-data methods relying on vision-text models such as CLIP \citep{clip, compbench, ghosh}. %We measure reachability by the proportion of images that when implementing the method of prompting, steering on the prompt space or steering on the $h$-space, successfully contain the target concepts.
To evaluate the concepts within the generated images, we train three classifiers for identifying the back shape, front shape and back-front colour pairs of the two shapes in each image. The resulting labels are then compared against the target concepts. Images producing incomplete outcomes (such as a black background with no shapes or an image only showing one shape) are accounted for as incorrect images (see Appendix \ref{app:classifiers}). Throughout our experiments, we train four models initialised with different random seeds and report the mean results obtained. %For the implementation of steering on the prompt space and $h$-space, we compare the reachability from different starting prompts $y_s$ to the target concepts. For example, for the target combination $(red, triangle, green, square)$, we compare the reachability of implementing steering from prompts such as ``a red triangle behind a green square" (semantic information and target concepts are the same) to ``a red circle behind a green square" or ``a blue triangle behind a red square" (several concepts are different to the target concepts). This allows for varying levels of difficulty in the steering of $y_s$, and provides insight on the ability of each of the steering methods to successfully reach concepts.

\section{Empirical Results} \label{sec:reachability}

%Please add below experimental findings that are \done{done}, \inprogress{in progress}, or are \todo{to be discovered} in the appropriate analysis categories. Also add links to assets (notebooks, figures, etc.) that are relevant to those findings. Make the findings concise!

\subsection{Baseline: Balanced Dataset} \label{sec:baseline}

\begin{figure}[htbp]
    \centering
    \includegraphics[clip, trim=0 0 0 0, width=0.9\columnwidth]{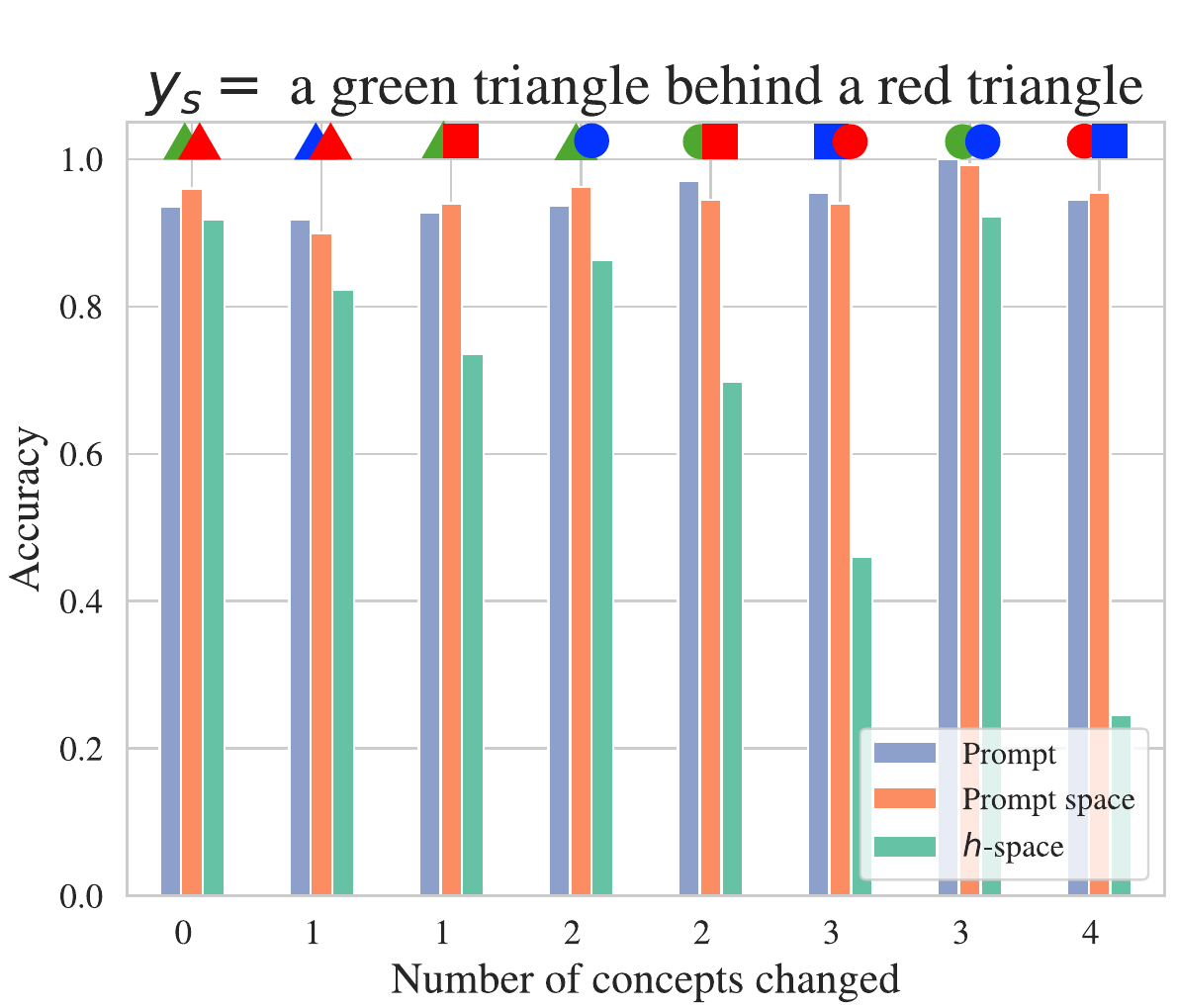}
    \caption{Reachability to different concept combinations when prompting, as well as steering from the starting prompt $y_s = $``a green triangle behind a red triangle". Target colour-shape combinations (note that the relative position is representative) are shown at the top of each bar, and are organised according to the number of concepts that differ from the concepts of $y_s$.} %Results are averaged across four models trained on the same dataset with different random seeds.} %The images $\mathcal{Z}$ containing the target concept were generated synthetically using the same code as the train set images $\mathcal{X}$.
    \label{fig:exp1-methods}
\end{figure}

Before introducing additional complexities to the data, we establish a clear understanding of the different reachability methods under balanced conditions. Specifically, we train diffusion models and use steering to generate various target combinations, varying the level of complexity between these and the starting prompt. Reachability when directly prompting the target combination is also accounted for, as illustrated in Figure \ref{fig:exp1-methods}.

\paragraph{Prompt-space steering matches prompting reachability} Prompting and steering on the prompt space yield an accuracy above 0.9, demonstrating their effectiveness in reaching concepts under balanced conditions. This highlights steering as a reliable method for accessing desired concepts. In contrast, steering within the $h$-space exhibits a pronounced dependency on the targeted concepts (for fixed $y_s$). Notably, accuracy on the $h$-space diminishes as the number of modified concepts increases, suggesting a limitation in its capacity to handle heavily complex modifications.

\paragraph{Reachability bias when steering on the $h$-space} We observe a tendency for reachability to be higher, relative to other concept combinations with the same number of modified concepts, when the target front shape and back shape are identical (second column from the right of Figure \ref{fig:exp1-methods}). We identify this as a property of the dataset, as it is consistently observed across different models, and find it aligns with the steerability bias reported by \citet{steerability}. This supports the role of dataset properties in shaping reachability behaviour. We provide further analysis of reachability on the $h$-space in Appendix \ref{app:norm}.

Overall, we observe that prompting and steering perform effectively in a perfectly balanced dataset. In the following sections, we present results regarding their performance under adverse data conditions. Based on the results observed under balanced conditions, we choose the starting prompt that maximises performance when steering on the $h$-space. Unless stated otherwise, throughout the remaining experiments we choose to implement steering from the starting prompts $y_s$ which describe the target concepts (0 concepts changed).
%however, as shown in the sections below, it is highly sensitive to any form of corruption in the data. This observation raises a key question: under what conditions does prompting fail, and do such failures imply that the corresponding concepts become entirely unreachable?

\subsection{Scarcity of Concepts} \label{sec:number}

%We investigate how the presence of a concept in the dataset $\mathcal{X} \times \mathcal{Y}$ will impact reachability. In particular, 
We fix  $(red, triangle, green, square)$ as a target concept combination and gradually reduce the number of data points in the subset $[c_1 = red]_\mathcal{X}$. This process is repeated for the subsets $[s_1 = triangle]_\mathcal{X}$, $[c_2 = green]_\mathcal{X}$, and $[s_2 = square]_\mathcal{X}$. We present two of these tests in this section and relay the remaining cases to Appendix \ref{app:number}.

\begin{figure}[htbp]
    \centering
    \includegraphics[width=1\columnwidth]{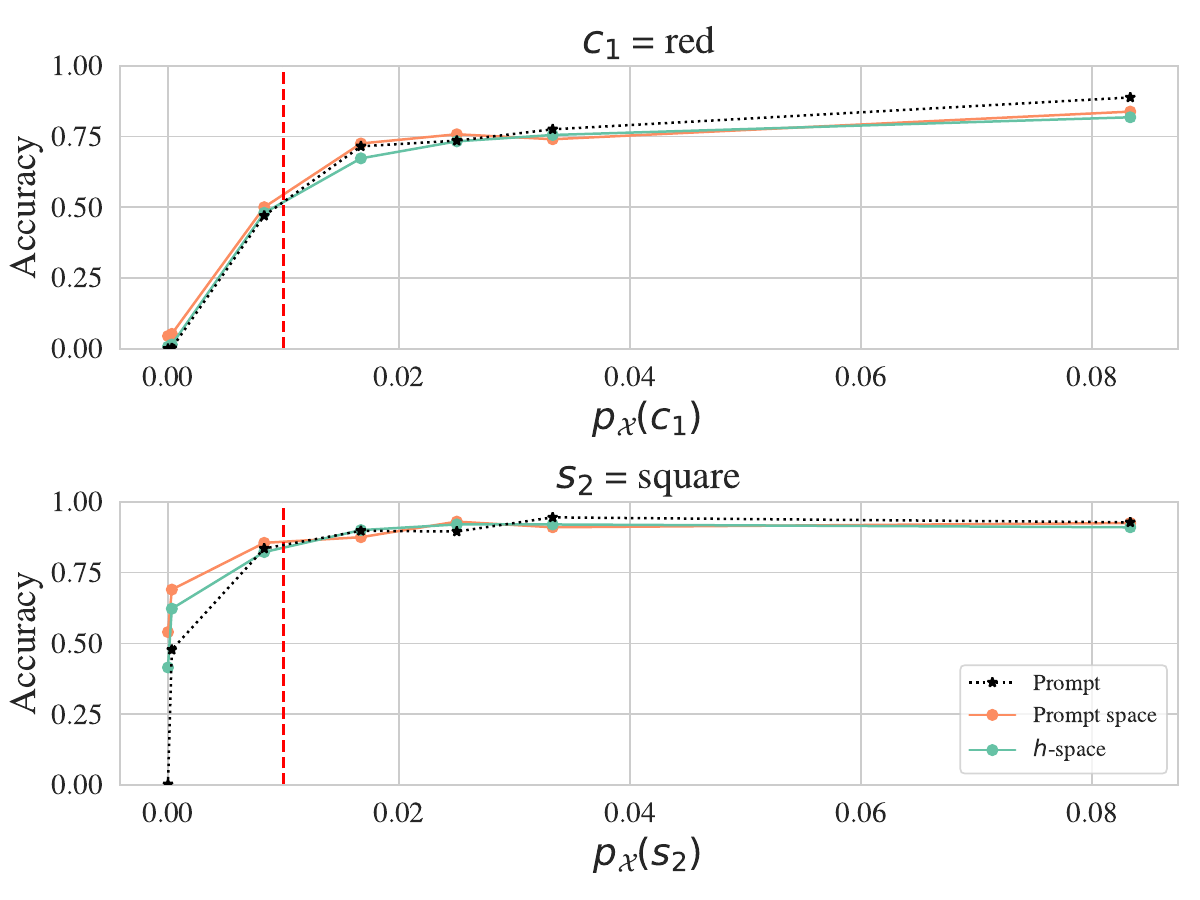}
    \caption{Accuracy of prompting and steering on the prompt space and $h$-space to $(red, triangle, green, square)$ for starting prompt $y_s = $ ``a red triangle behind a green square" and varying proportion $p_\mathcal{X}$ of images in the dataset containing the concepts $c_1 = red$ (top) and $s_2 = square$ (bottom) across the dataset. %Images containing the target combination were generated through a diffusion model trained on the balanced dataset.
    The dashed red line marks the approximate threshold 0.01 of the shift in reachability.}
    \label{fig:exp2-threshold}
\end{figure}

\paragraph{Reachability drops sharply past a critical threshold} We implement steering to our chosen target concept combination and observe that, although reachability decreases as the presence of concepts is reduced, this decline remains gradual until reaching a threshold at $p_\mathcal{X}(f) \approx 0.01, f = c_1, s_2$. Below this, reachability decreases rapidly (Figure \ref{fig:exp2-threshold}). This decline is less pronounced for steering, most noticeably for $s_2 = square$, highlighting that \emph{steering can be more effective at maintaining concept reachability under conditions of scarcity.} This behaviour suggests a phase transition-like effect: when concept presence falls below $1\%$, the system shifts abruptly from a state of high reachability to one of significantly diminished reachability. This phase transition implies that the model requires few data points, relative to the train set size, in order to learn a concept effectively. Increasing the number of data points containing a concept beyond this critical threshold has limited impact on improving reachability.

\paragraph{Positionally OOD combinations can be reachable through steering}
When the presence of a concept $f$ is such that $p_\mathcal{X}(f) = 0$, generating the combination $(red, triangle, green, square)$ results in a positionally OOD target. For instance, eliminating the data points in $[c_1 = red]_\mathcal{X}$, limits the model's exposure to red shapes in the back position, even if red shapes remain present in the front position $c_2$. Our findings reveal that models are unable to reach the target concept combination through prompting (Figure \ref{fig:exp2-threshold}), aligning with observations by \citet{mouse_chase}. In contrast, steering methods can achieve moderate reachability. Notably, when reducing $[s_2 = square]_\mathcal{X}$, steering on the prompt space achieves more than 50\% accuracy, even when no training images contain the reduced concept. This indicates that when prompting fails, alternative mechanisms may enable models to generalise concepts to unseen positions.

%As the presence of a concept in one position is reduced, we examine the model's capacity to generalise to the missing position. 

\subsection{Underspecification of Captions} \label{sec:specification}

We reduce the factors of interest of the dataset, initially $(c_1, s_1, c_2, s_2)$. For instance, removing the specification of $c_1$ results in captions $y$ containing information only with respect to $(s_1, c_2, s_2)$. See Appendix \ref{app:labelling} for details. We then evaluate the reachability of $(c_1, s_1, c_2, s_2)$ via prompting and steering.

We present the results of prompting when describing the full set of target concepts. For example, ``a red triangle behind a green square". Steering is performed on top of starting prompts $y_s$ that reflect only the concepts seen during training. For example, when removing back colour ($c_1$), steering is implemented from $y_s = $ ``a triangle behind a green square" to images with the concept combination $(red, triangle, green, square)$.

%\begin{figure*}[tb]
%    \centering
%    \includegraphics[width=1\textwidth]{main_plots/exp3/image.png}
%    \caption{a) Average reachability across 10 randomly chosen concept combinations $(c_1, s_1, c_2, s_2)$ for different levels of label specification. %The label information for low specification is chosen to remove information on the background shape of the images in the train set. 
%    b) Outputs produced by a model when steering towards $(red, circle, green, square)$ when the factors $c_1$ (left) and $s_2$ (right) are removed from the captions of the train set. Each diagram shows the target concept in the top axis, and the alternative concepts the removed factor can take in the remaining directions. The label $X$ represents any other output, including those that do not produce the target concept for the seen factors. Each diagram displays the proportion of images of the output belonging to each label. A reachability method that correctly generates the target concept will produce a high proportion of images on the top axis. If a model only fails to generate the removed concept value, the proportion of images at $X$ will be low.}
%    \label{fig:exp3-reachability}
%\end{figure*}

\begin{figure*}[htbp]
    \centering
    \begin{subfigure}{0.42\textwidth}
        \includegraphics[width=\linewidth]{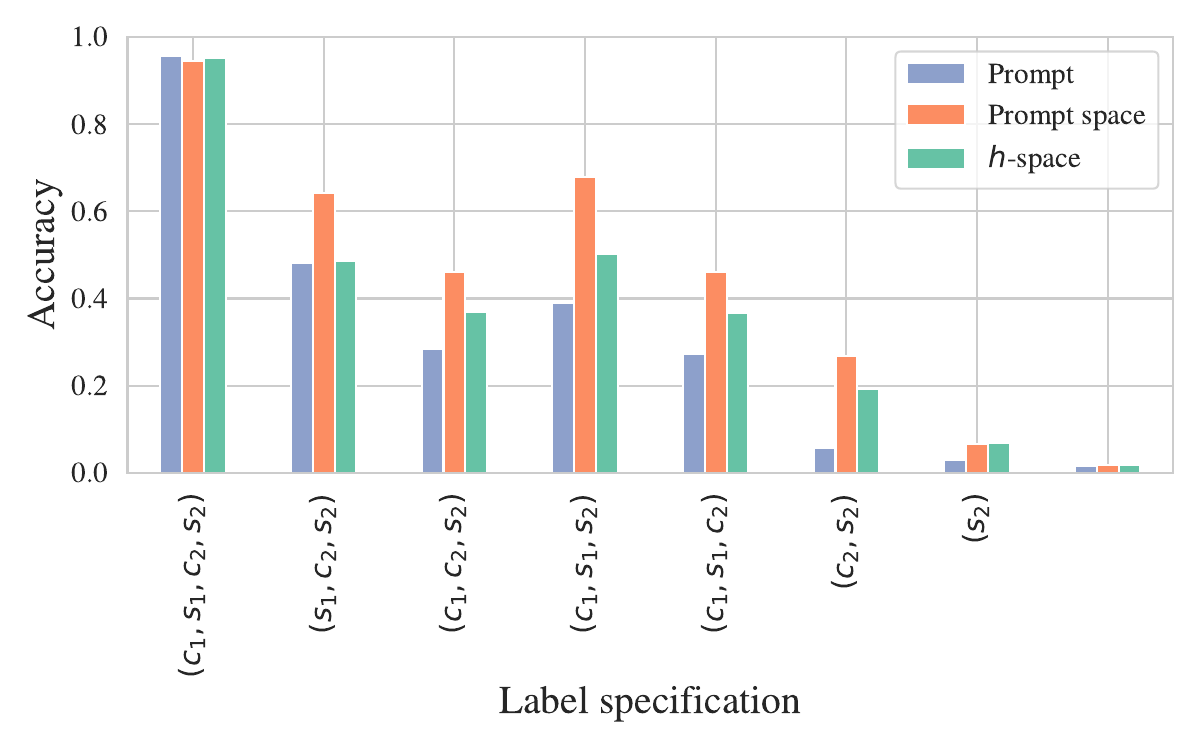}
        \caption{} % No individual caption
        \label{fig:exp3-reachability}
    \end{subfigure}
    \hfill
    \begin{subfigure}{0.55\textwidth}
        \includegraphics[width=\linewidth]{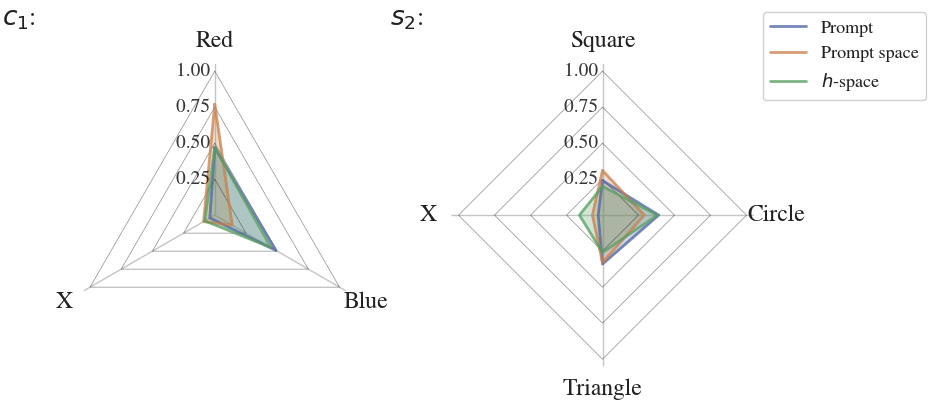}
        \caption{} % No individual caption
        \label{fig:exp3-reachability-b}
    \end{subfigure}
    \caption{a) Average reachability across 10 randomly chosen concept combinations $(c_1, s_1, c_2, s_2)$ for different levels of label specification. %The label information for low specification is chosen to remove information on the background shape of the images in the train set. 
    b) Outputs produced by a model when steering towards $(red, circle, green, square)$ when the factors $c_1$ (left) and $s_2$ (right) are removed from the captions of the train set. Each diagram shows the target concept in the top axis, and the alternative concepts the removed factor can take in the remaining directions. The label $X$ represents any other output, including those that do not produce the target concept for the seen factors. Each diagram displays the proportion of images of the output belonging to each label. A reachability method that correctly generates the target concept combination will produce a high proportion of images on the top axis. If a model only fails to generate the removed concept value, the proportion of images at $X$ will be low.}
    
\end{figure*}

\paragraph{Steering outperforms prompting under low specification}
In all cases of specification reduction, we observe an improvement in reachability with respect to prompting when using either steering method (Figure \ref{fig:exp3-reachability}). Among these, steering on the prompt space achieves the highest reachability, demonstrating it is a more effective approach for accessing target concepts under reduced specification conditions.

\paragraph{A decrease in specification hinders reachability} Despite the improvement in reachability achieved through steering, reducing the specification of captions during training results in a rapid decrease in reachability. Figure \ref{fig:exp3-reachability} shows average accuracy falls below $0.30$ for models trained with fewer than two concepts specified. Classifying the outputs after prompting and steering reveals that models tend to generate concepts accurately for the factors specified during training, but fail to correctly generate the missing concepts. An example for one random seed is presented in Figure \ref{fig:exp3-reachability-b}. When prompting is used, the model's accuracy on specified labels is high and on unspecified labels is nearly equivalent to chance-level performance. In contrast, steering yields higher accuracy on the full target concept combination, although remaining close to the prompting baseline. These findings show captions hold a pivotal role in organising and structuring a model's latent space, providing a semantic framework that enhances its ability to access concepts. Our results demonstrate that more detailed labeling of training images can improve the success rate of generating target concept combinations. On the contrary, given a trained model, overly detailed prompts may not lead to an improvement of reachability.

We further analyse reachability when using $y_s$ containing the full semantic information of the target combination $(c_1, s_1, c_2, s_2)$ in Appendix \ref{app:label_results}.

\subsection{Biases} \label{sec:tied_concepts}

\begin{figure}[htb]
    \centering
    \includegraphics[width=1\columnwidth]{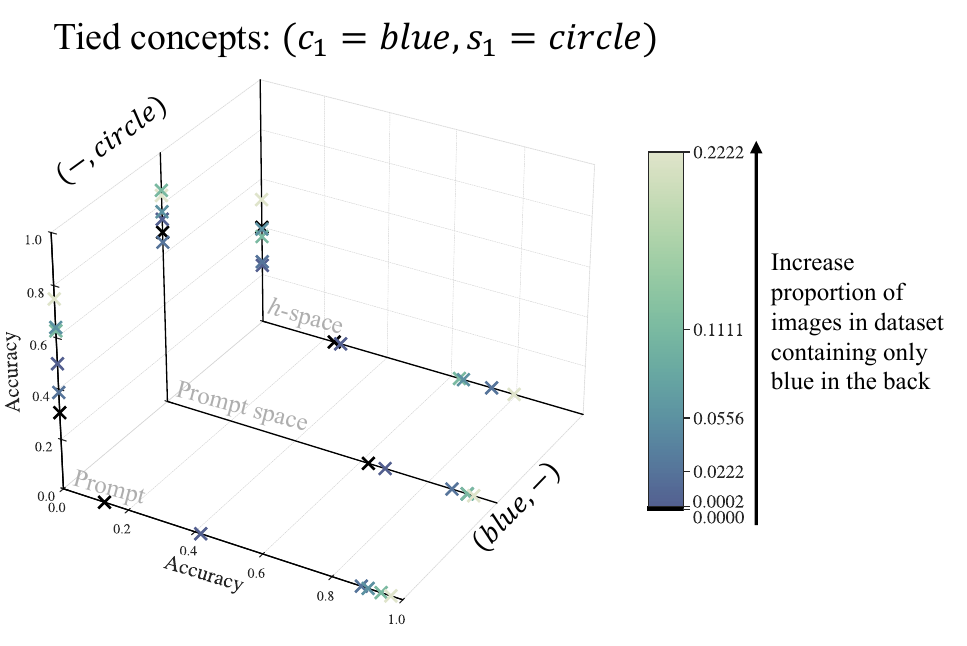}
    \caption{Reachability to concept combinations containing non-blue circles in the back (vertical axis) and blue non-circular shapes in the back (horizontal axis). Results are averaged over 6 randomly chosen target concept combinations, with starting prompt $y_s$ describing those concepts. The X's mark the accuracy of reaching either type of concept when prompting, steering on the prompt space and steering on the $h$-space. A lighter colour represents a stronger presence of images containing only blue in the back in the dataset.}
    \label{fig:exp4-start-end}
\end{figure}

We measure reachability to individual concepts that are consistently paired during training. Specifically, we construct a dataset where an image in the training set satisfies $c_1 = blue$ if and only if $s_1 = circle$. To evaluate the model's ability to disentangle these concepts, we measure reachability to concept combinations of the form $(blue, s_1^\prime, c_2, s_2)$, where $s_1^\prime \neq circle$, and $(c_1^\prime, circle, c_2, s_2)$, where $c_1^\prime \neq blue$. %By doing so, we analyse the model’s capability to disentangle and independently access each concept.
Prompt space steering exhibited minimal variation across different starting prompts $y_s$, whereas on the $h$-space, starting with a prompt describing the target concepts of interest yielded the highest accuracy. We present this in Figure \ref{fig:exp4-start-end}.

\paragraph{Biases in compositional OOD generalisation can be bypassed through steering} When no train images contain only blue in the back, attempting to reach combinations where either concept appears separately places the model in a compositionally OOD setting. Figure \ref{fig:exp4-start-end} shows steering, particularly on the prompt space, achieves a moderately high reachability to both concepts (accuracy of black X's), suggesting that models are capable, to some extent, of disentangling heavily biased concepts. Prompting fails to achieve this level of disentanglement, revealing its inability to accurately reflect a model’s full capabilities.

%Our results for the scenario where the two concepts studied are always tied in the train set show that regular prompting struggles to access each of the concepts separately. This is shown in Figure \ref{fig:exp4-start-end} (black cross markings). This setting involves compositional OOD generalisation, which produces a higher reachability (accuracy greater than 0.1) than the positional OOD generalisation we present in Section \ref{sec:number}, where accuracy was 0. However, reachability is significantly low compared to the accuracies observed by steering on the prompt space and $h$-space, showing that a failure of the model to produce the correct outcome by prompting is not due to the model's inability to reach the concept, but due to the failure of the prompt method to reach the concept properly.

\paragraph{Increasing the presence of an individual concept improves the separate reachability of both tied concepts} %Our results show that when there are few images containing only one of the two concepts (in this case, blue in the back), reaching either concept individually by steering is more successful than prompting.
%As the number of images containing blue shapes (not circles) is increased, the reachability of blue in the back increases rapidly, as shown in the gap of values between 0.6 and 0.8 on the horizontal plane in Figure \ref{fig:exp4-start-end}, with a similar threshold pattern as observed in Section \ref{sec:number}. Moreover, the reachability of the other tied concept (non-blue circles in the back), although with some minor variability, also experiences an increase in reachability. This pattern implies that the model learns to identify the contrasting concept, likely by inferring its definition through what it is not.
As the number of images containing non-circular blue shapes increases, the reachability to combinations only containing $c_1 = blue$ rises rapidly, evidenced by the large increase in accuracy along the horizontal axis in Figure \ref{fig:exp4-start-end}. This shows a similar threshold pattern to that observed in Section \ref{sec:number}. Interestingly, the lighter X's on the vertical axis tend to achieve the highest accuracy, suggesting that the reachability of non-blue circles in the back also increases, albeit with minor variability. This pattern implies that the model learns to identify the contrasting concept, likely by inferring its definition through what it is not.
%, demonstrating an emergent capacity for relational understanding and differentiation between the two concepts.

See Appendix \ref{app:tied_concepts} for additional results.

\subsection{Reachability in Real Settings}
\begin{figure}[htb]
    \centering
    \includegraphics[width=1.0\columnwidth]{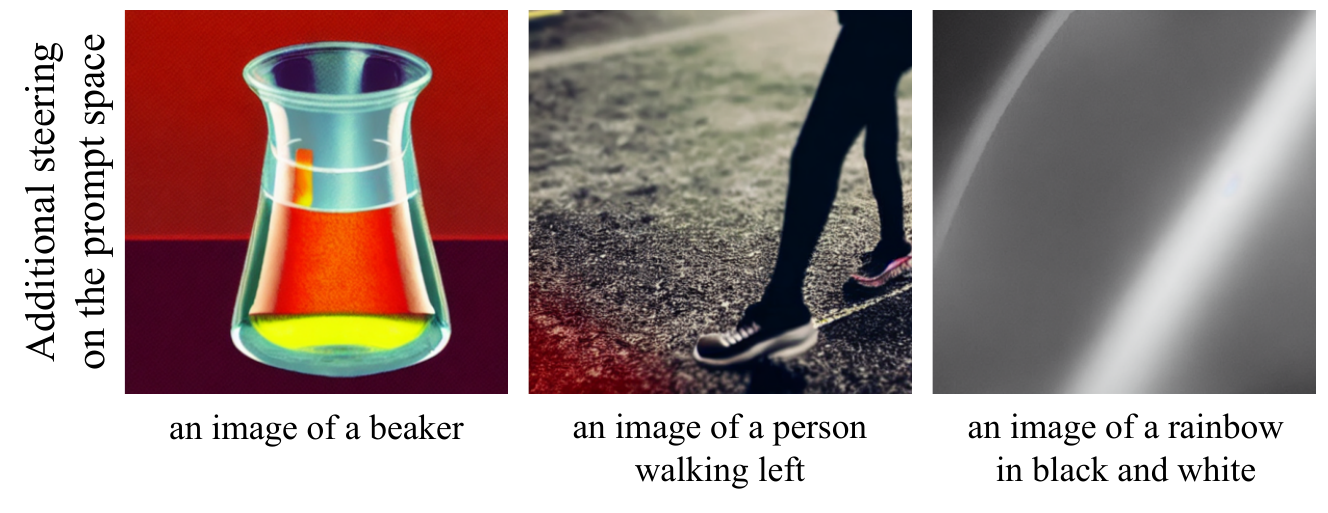}
    \caption{Images sampled from the same random seeds used in Figure \ref{fig:stable_diff} with steering vectors added to the prompt space during sampling. Using steering, we observe: (L) a more accurate representation of a beaker, (C) a successful change in orientation of the person towards the left, (R) the rainbow arc generated in greyscale tones.}
    \label{fig:stable_diff_steer}
\end{figure}

We additionally implement steering on the prompt space of Stable Diffusion in order to explore how reachability is impacted in a real-world setting. Figure \ref{fig:stable_diff_steer} presents samples from the same initial random seed $\mathbf{x}_T$ as in Figure \ref{fig:stable_diff}, with additional steering implemented. The resulting images show a clear improvement, more accurately representing the target concepts. Further examples and analysis of reachability through steering are provided in Appendix \ref{app:stable_steer_prompt}. Overall, we observe that steering can extend beyond the reachability of prompting. However, dataset limitations such as underspecification continue to hinder reachability, exhibiting patterns consistent with those observed in our synthetic analysis. %We highlight the need to further explore reachability in more complex, controlled datasets to gain deeper insights into real-world models.
%Overall, we observe that steering can improve outcomes, as shown in Figure \ref{fig:stable_diff_steer}, but that there exist limitations which may require further exploration of reachability methods, as well as further analysis of the properties of highly complex latent spaces.

Furthermore, we conduct a similar analysis to that in Section \ref{sec:reachability} using a subset of the CelebA dataset. Results are presented in Appendix \ref{app:celeba}. Specifically, we use the labelled attributes for Gender (male/female) and Hat (wearing/not wearing a hat) as the concepts to be analysed. Overall, we again find that the structure of the training data strongly influences concept reachability. In particular, reachability declines sharply when the number of training examples containing a given concept falls below a certain threshold, and underspecified labels substantially hinder access to those concepts. Additionally, we observe that reducing biases in the co-occurrence of concepts improves a model’s ability to represent and generate individual concepts in isolation. These findings are consistent with those observed in the synthetic setting, supporting the generality of these conclusions and highlighting their relevance in real settings.

\section{Discussion and Conclusion}

This paper explores the limits of reachability in diffusion models, examining the challenges in accessing target concepts. While prompting is the default approach, our findings reveal that it often fails to fully capture a model’s potential for generating specific concepts. We demonstrate that steering can enhance the reachability of scarce concepts and disentangle biases inherent in datasets. We also identify key patterns in reachability such as shifts in a model's ability to access concepts and the critical role of proper concept specification in captions. This work provides a valuable framework for evaluating when concepts are reachable, and offers insights into how we can design train sets to improve the generative performance of diffusion models.
%We study three synthetics tasks that aim to capture the nature of problems in real data. 
%We conclude that (1) there is a threshold after which increasing the presence of a concept across a dataset does not significantly affect its reachability, (2) adding information to prompts beyond the specifications seen during training does not improve reachability, and (3) disentangling tied concepts in the train data by increasing the data points containing individual concepts can be beneficial for the model's reachability to either concept. Our results provide insight into how train sets can be improved to reduce mismatches between a model's output and a target output.

The improvement observed in reachability through steering suggests a misalignment between the model's semantic and visual understanding of concepts. That is, a model may be able to reach a concept, but the prompt fails to access the correct spatial information required to generate it. Moreover, we observe differences in the accuracy of steering on the prompt space and steering on the $h$-space, which highlight the importance of the stage of the transformation at which steering is performed. In particular, the nature of the space in which steering is implemented will impact reachability. For example, while the text encoding is independent of the timestep $t$, the bottleneck output ($h$-space) depends on it. The two spaces also differ in dimensionality, and the degree of disentanglement of either space may also influence steering effectiveness. We note that, while steering on the $h$-space does not perform optimally for every choice of $y_s$, our results show it is an effective method when $y_s$ aligns with the target concept combination. 

An intriguing aspect of our experiments concerns a model's ability to employ different mechanisms (compositional or positional) to achieve OOD generalisation. As presented in Section \ref{sec:number}, models face significant challenges in OOD positional generalisation, with instances where reachability across all methods approaches zero. While biased compositional OOD generalisation remains a demanding task, we observe moderate levels of reachability using at least one of the methods studied (Section \ref{sec:tied_concepts}). Generalising shape positionally from back to front is particularly interesting, as it requires the model to synthesise information from multiple images to reconstruct a complete, unobstructed shape. We detect no substantial variance in reachability when positionally generalising in either direction, suggesting that models find both tasks equally demanding. 

Despite improvements in reachability, the steering methods explored in this study rely on an auxiliary collection of images $\mathcal{Z}$ containing the target concept combinations, which in real data may be hard to acquire. Investigating alternative steering methods that mitigate this dependency and enhance reachability remains an important direction for future research. Additionally, there may exist simple or more efficient alternative methods to steering that outperform existing approaches and warrant further exploration. Moreover, while our work isolates the impact of each obstacle studied, in real data it is likely that problems occur simultaneously and with varying degrees of complexity. Expanding our dataset to incorporate additional factors and more intricate relationships can provide a more comprehensive analysis of model behaviour.

\section*{Acknowledgements}
MAR is supported by the Department of Mathematics, Imperial College London, through the Roth Scholarship.

\section*{Impact Statement}

This paper focuses on studying the robustness methods for model control. It holds direct value for model providers as it highlights the delicacy of prompting and instead motivates the usage of alternative latent control methods.

\newpage

\bibliographystyle{icml2025}
\bibliography{references_lit_review.bib}

\newpage
\appendix

\onecolumn

\section{Hyperparameters} \label{app:hyperparameters}
\subsection{Architecture}

For our experiments, we used the Diffusers package \citep{diffusers} to implement the U-net architecture. Our model comprises 3.7 million parameters and features a symmetric architecture with four downsampling and upsampling blocks. The channels in each block are set to $(16, 32, 64, 128)$, respectively. Each block contains two ResNet blocks.

Additionally, we incorporate cross-attention layers in the midblock, enabling the model to integrate information from a text-encoding input. For processing the text prompt, we use a pre-trained T5Small text encoder \citep{t5}. The encoder's weights are frozen throughout training of the U-net and during the optimisation of concepts vectors.

\subsection{Training}

Training of the U-net is performed for 70 epochs using Adam with learning rate $0.001$ and default parameter values. Additionally, we use an exponential learning rate scheduler with parameter $gamma = 0.98$. All models are trained using $T=1000$ and sampled with a DDPMScheduler at inference time.

\subsection{Concept Vector Optimisation}

\begin{figure}[htb]
    \centering
    \includegraphics[width=1\columnwidth]{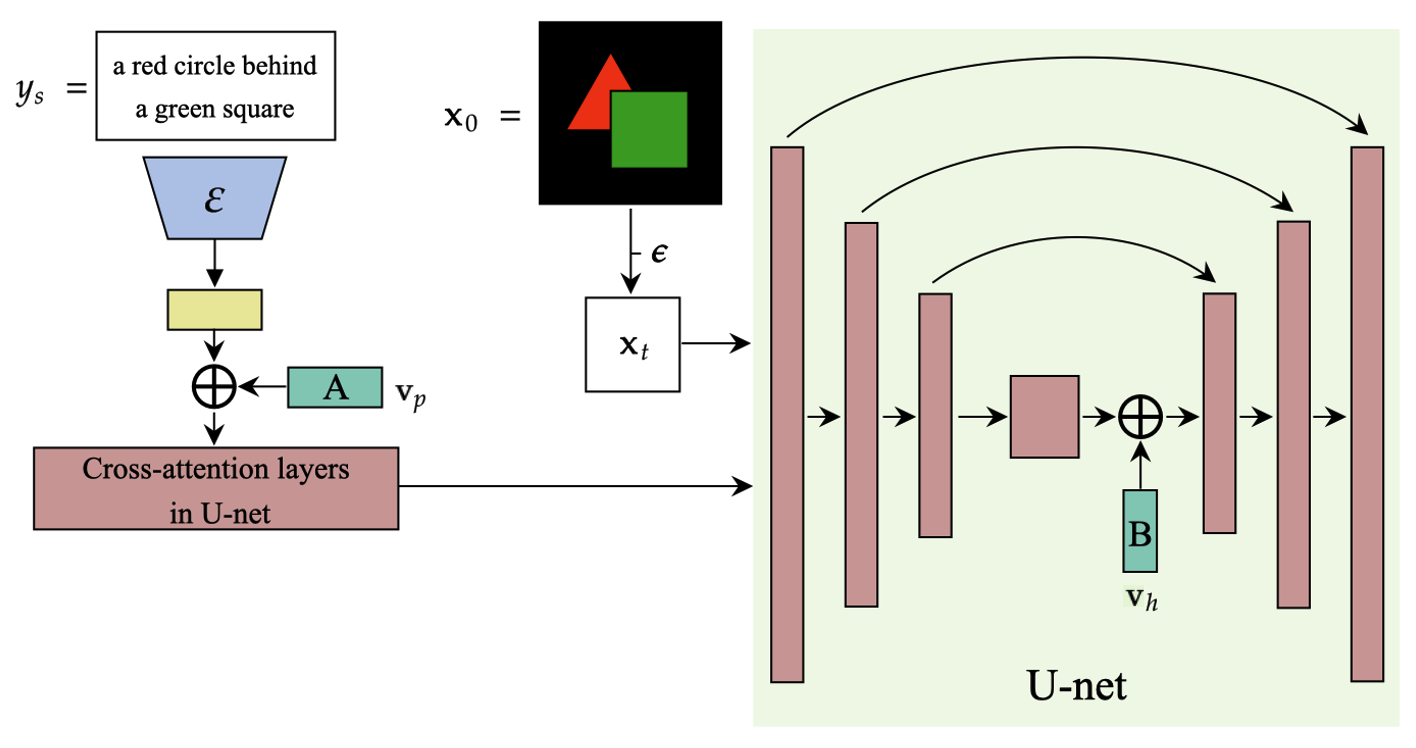}
    \caption{Diagram of spaces in the architecture where steering is implemented. A: prompt space, the concept vector is added to the encoding for the text prompt before passing through the cross-attention layers of the U-net. B: $h$-space, the concept vector is added to the bottleneck layer of the U-net, after the mid-block.}
    \label{fig:steering_unet}
\end{figure}

Concept vectors are initialised at the zero-vector, and optimised for 5000 steps using Adam, with learning rate 0.02 and default parameter values. This choice is observed to be sufficient for the optimised losses $L_p$ and $L_c$ to converge. A diagram of the different spaces in which the optimisation is implemented is shown in Figure \ref{fig:steering_unet}.

For steering on the prompt-space, $\mathbf{v}_p$ is of the same dimensionality as the input prompt $y_s$. In the original dataset with captions ``a $\{c_1\}$ $\{s_1\}$ behind a $\{c_2\}$ $\{s_2\}$, this accounts for $\mathbf{v}_p \in \mathbb{R}^{1\times 10 \times 512}$, and is adjusted for the experiments in Section \ref{sec:specification}.

The dimension of the vectors $\mathbf{v}_h$ optimised in the bottleneck of the U-net is $\mathbf{v}_h \in \mathbb{R}^{128 \times 8 \times 8}$ for all experiments.

\section{Dataset Creation} \label{app:data}

\begin{figure}[htbp]
    \centering
    \includegraphics[width=0.8\columnwidth]{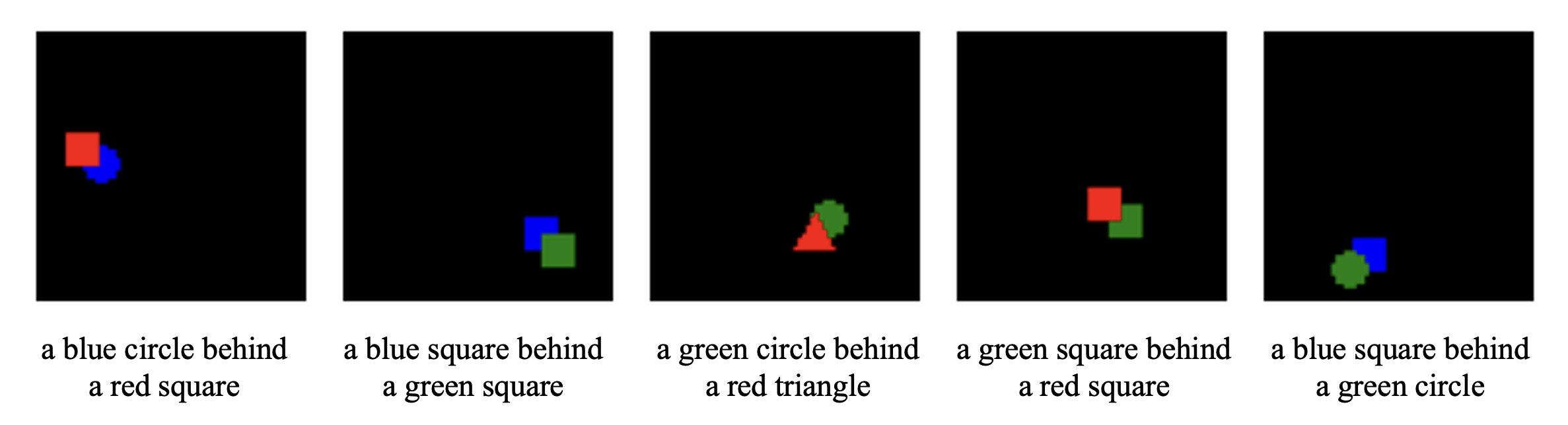}
    \caption{Examples of image-caption pairs from our synthetic dataset. Each image shows two shapes of different colours, one partially covered by the other, on a black background.}
    \label{fig:trainset}
\end{figure}

Images are of dimension $64 \times 64$ pixels, comprised of two coloured shapes. To create the complete dataset, we generate 1000 images with each of the combinations of the concepts of interest $(c_1, s_1, c_2, s_2)$. Each image is created using the Pillow package in Python \citep{pillow}. First, the location of the centre of the back shape is determined, and then the position of the front shape is sampled uniformly from a neighbourhood of this centre. This neighbourhood ensured a reasonable portion of the back shape was always visible for every pair of shapes (thus, facilitating the classification task). The minimum percentage of each back shape visible in the balanced train set for the diffusion models is the following: circle 52.46\%, triangle 48.97\%, square 59.38\%. Examples of train images are presented in Figure \ref{fig:trainset}.

%Assuming all shapes are bounded by a circle of radius $r = 5$ pixels, the following steps were taken to generate an image:
%\begin{enumerate}
%\item The location of the back shape in each image was sampled uniformly across the $(x_b, y_b) \in [2r, 64 - 2r] \times [2r, 64 - 2r]$ plane.
%\item The location of the front shape was sampled uniformly, subject to two constraints:
%\begin{itemize}
%\item It must be within a specified distance from the center of the back shape to ensure overlapping of the shapes, $x_f \in [x_b - 0.9 r, x_b + 0.9 r]$, $y_f \in [y_b - 0.9 r, y_b + 0.9 r]$
%\item It must not overlap too closely with the center of the back shape, ensuring that the back shape remains identifiable, $d((x_f, y_f), (x_b, y_b)) > r$.
%\end{itemize}
%\end{enumerate}

When reducing the size of some of the combinations of the dataset, the size of the remaining combinations is proportionally increased to ensure that the total dataset size is as close as possible to 54,000 while maintaining uniformity (equal number of images) among the unaffected concept combinations.

\section{Classifier and Evaluation Details} \label{app:classifiers}

We train three different classifiers for identifying the back shape, front shape and back-front colour pairs. The architecture consists of two convolutional layers that increase channel size to 16 and 32 respectively. Both consist of $3\times3$ kernels with ReLU activation and $2\times2$ max-pooling. The flattened output is passed to a fully connected layer of 128 units and ReLU. The output vector is three-dimensional in the case of both shape classifiers, and consists of six dimensions for the classifier of back-front colour pairs.

Each classifier is trained using Adam with learning rate $0.001$ on a dataset comprised of the original 54,000 balanced dataset and an equal number of sampled images from trained diffusion models. Training is implemented for 7 epochs. %, with models achieving accuracy higher than $99\%$ on a test set of 10,000 images. 
When evaluating on 5400 human-labelled images that were generated by prompting from two additional diffusion models trained on balanced data, the classifier obtains an accuracy of 96.63\%.

During evaluation of reachability, 100 images are sampled using a reachability method. Each of these is labelled according to the outputs of the trained classifiers. Images are accounted for as correct if the labels match the target concept combination. In cases where classifier results were uncertain or ambiguous, additional human evaluations were conducted. Furthermore, we identify the number of (non-black) colours in an image, and if this number if distinct from 2, we account for the images as incorrect. After this, the proportion of correct images is used as the reachability value.

\section{Reduction of Label Specification on Captions} \label{app:labelling}

Our dataset is originally comprised of captions of the form ``a \{$c_1$\} \{$s_1$\} behind a \{$c_2$\} \{$s_2$\}". In order to reduce the semantic information of the prompt, we implement the following:

\begin{enumerate}
    \item Remove $c_1$: replace the caption with ``a \{$s_1$\} behind a \{$c_2$\} \{$s_2$\}"
    \item Remove $s_1$: replace the caption with ``a \{$c_1$\} shape behind a \{$c_2$\} \{$s_2$\}"
    \item Remove $c_2$: replace the caption with ``a \{$c_1$\} \{$s_2$\} behind a \{$s_2$\}"
    \item Remove $s_2$: replace the caption with ``a \{$c_1$\} \{$s_2$\} behind a \{$c_2$\} shape"
    \item Remove $c_1$ and $s_1$: replace the caption with ``a \{$c_2$\} \{$s_2$\}"
    \item Remove $c_1$, $s_1$ and $c_2$: replace the caption with ``a \{$s_2$\}"
    \item Remove $c_1$, $s_1$, $c_2$ and $s_2$: replace the caption with the empty string, ``"
\end{enumerate}

\section{Additional Experiments}

%\subsubsection{Attention} \label{app:attention}

%In this section we present additional heatmaps to the ones presented in Section \ref{sec:reachability}. In Figure \ref{fig:exp1-attention2}, we show an example of the entanglement of colour to the shapes being sampled in an image. The starting prompt is chosen to be $y_s = $``a green triangle behind a red triangle", and the target concept combination involves red triangles behind green triangles, reversing the outputted relation. We observe that the the shape that each of the tokens for ``triangle" (back and front) attends to is independent of the position the semantic information relates to, but to the colour of the shape being samples. Thus, when steering on the prompt space successfully modifies the output and reverses the positions of the shapes, the model attends to the shapes in the images, but in the wrong position described in the prompt.

%When the colour of a shapes is steered to a colour not mentioned in the prompt, the attention of that shape is reduced, as is the case of the blue triangle generated in Figure \ref{fig:exp1-attention}.

%\begin{figure}[htb]
%    \centering
%    \includegraphics[width=0.7\textwidth]{main_plots/exp1/attention_2.png}
%    \caption{Attention heatmaps for an image sampled from the prompt ``a green triangle behind a red triangle" when no steering, steering on the prompt space and steering on the $h$-space is implemented. The diagrams show a successful steering on the prompt space and unsuccessful steering on the $h$-space.}
%    \label{fig:exp1-attention2}
%\end{figure}

\subsection{Additional Analysis on a Balanced Data} \label{app:norm}

\begin{figure}[htbp]
    \centering
    \includegraphics[width=0.75\textwidth]{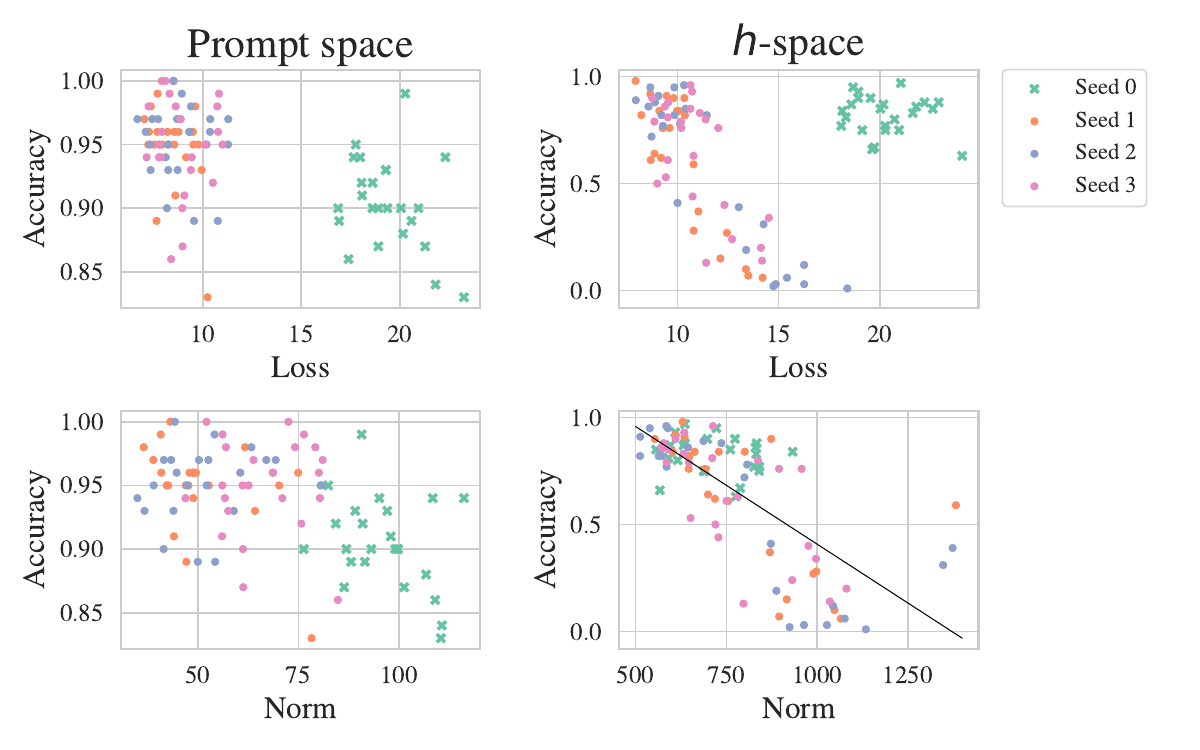}
    \caption{Relation between accuracy of steering on the prompt space and $h$-space and final loss (top row) and final vector norm (bottom row) after the optimisation of the concept vector. The different colours show the results for four different models trained on different random seeds.}
    \label{fig:exp1-norm_loss}
\end{figure}

\paragraph{Final norm of optimised concept vector is indicative of reachability when steering on the $h$-space} 
The differences in the accuracies achieved through steering on the $h$-space suggest that, from a given starting prompt $y_s$, certain concept combinations are inherently more reachable than others. 
While the final loss at the end of the optimisation of the steering vector might be considered a natural indicator of reachability via steering (lower loss associated with higher reachability), we observe cases where high final loss values correspond to high steering accuracies (Figure \ref{fig:exp1-norm_loss}, teal x's). Our findings indicate that the final vector norm provides a more reliable indicator, with a larger norm being associated with lower reachability (Figure \ref{fig:exp1-norm_loss}, bottom right). Interestingly, when sampling, adding a larger concept vector norm indicates a greater deviation from the latents that would be obtained from solely prompting $y_s$. Hence, we conclude that reachable concepts remain close in the $h$-space. Steering on the prompt space does not exhibit such patterns (Figure \ref{fig:exp1-norm_loss}, left). 

\subsection{Reachability When Removing One Concept Combination from the Dataset}

%In real datasets, it is unlikely that all possible combinations of concepts are present in the train dataset. Previous work has shown that models can compose concepts seen individually during training into a new, unseen combination.

\begin{figure}[htbp]
    \centering
    \includegraphics[width=1\columnwidth]{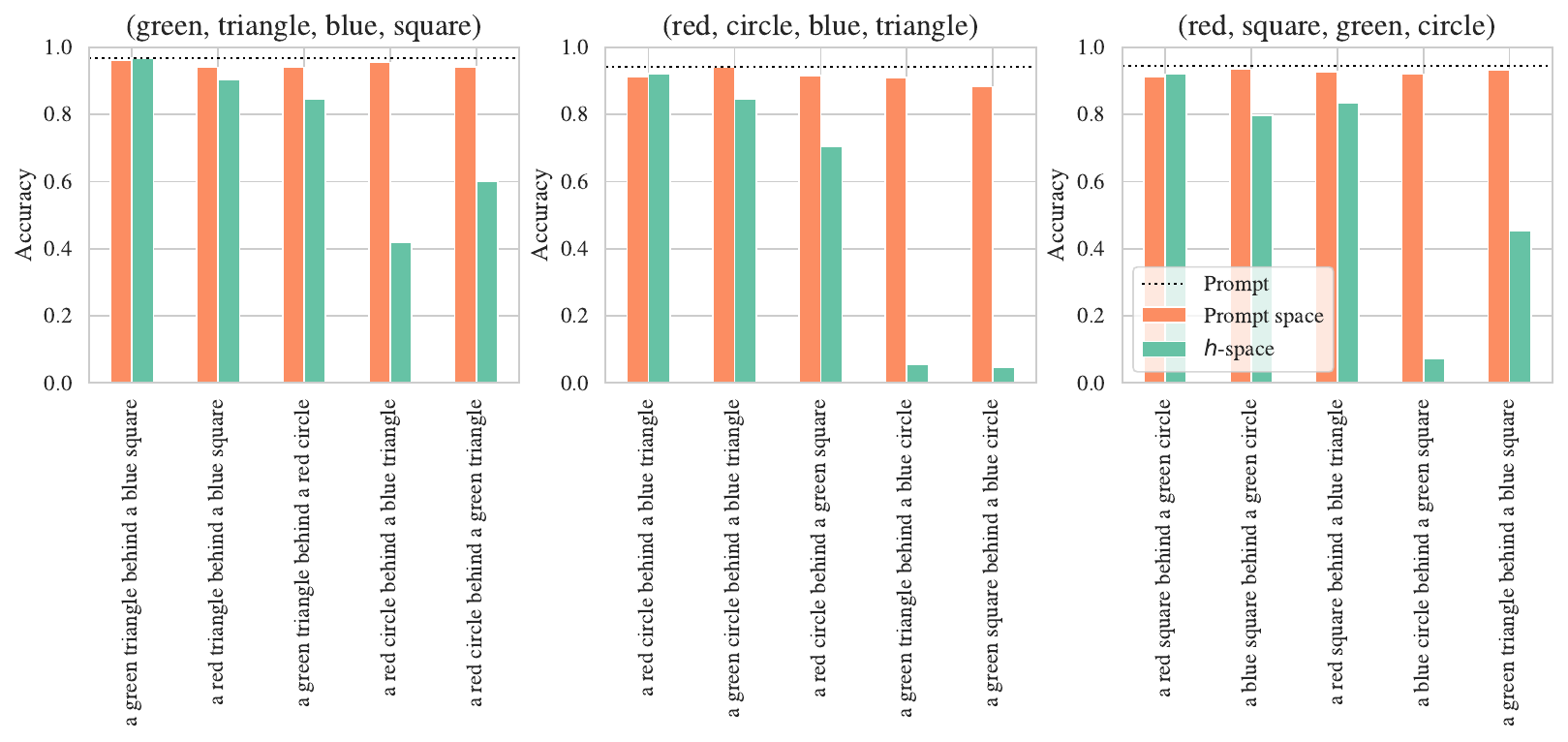}
    \caption{Steering accuracies for different starting prompts $y_s$ to $(green, triangle, blue, square)$, $(red, circle, blue, triangle)$ and $(red, square, green, circle)$ respectively. For each graph, the target concept combination is the only concept combination not present in the training dataset.}
    \label{fig:exp1-ood-simple}
\end{figure}

We test the ability of a model to generalise compositionally OOD in the simplest scenario, where only the target concept combination is removed from the dataset. Three examples of this are presented in Figure \ref{fig:exp1-ood-simple}. Our results highlight the variability of accuracy of steering on the $h$-space, also observed in Section \ref{sec:baseline}. Moreover, we observe that reachability is generally high and unaffected from the removal of the target combination. Prompting achieves the highest reachability, with steering on the prompt space also providing similar accuracy values. Thus, models are capable of generalising OOD for simple compositional concept combinations based on their ability to recombine learned individual concepts. We note that this is not necessarily extensible to any compositional OOD generalisation, as our experiments in Section \ref{sec:tied_concepts} provide a more complex compositional OOD generalisation task, where models are affected by the biases of combinations seen during training. 

\subsection{Additional Results for Scarcity of Concepts} \label{app:number}

\begin{figure}[htbp]
    \centering
    \includegraphics[width=1\columnwidth]{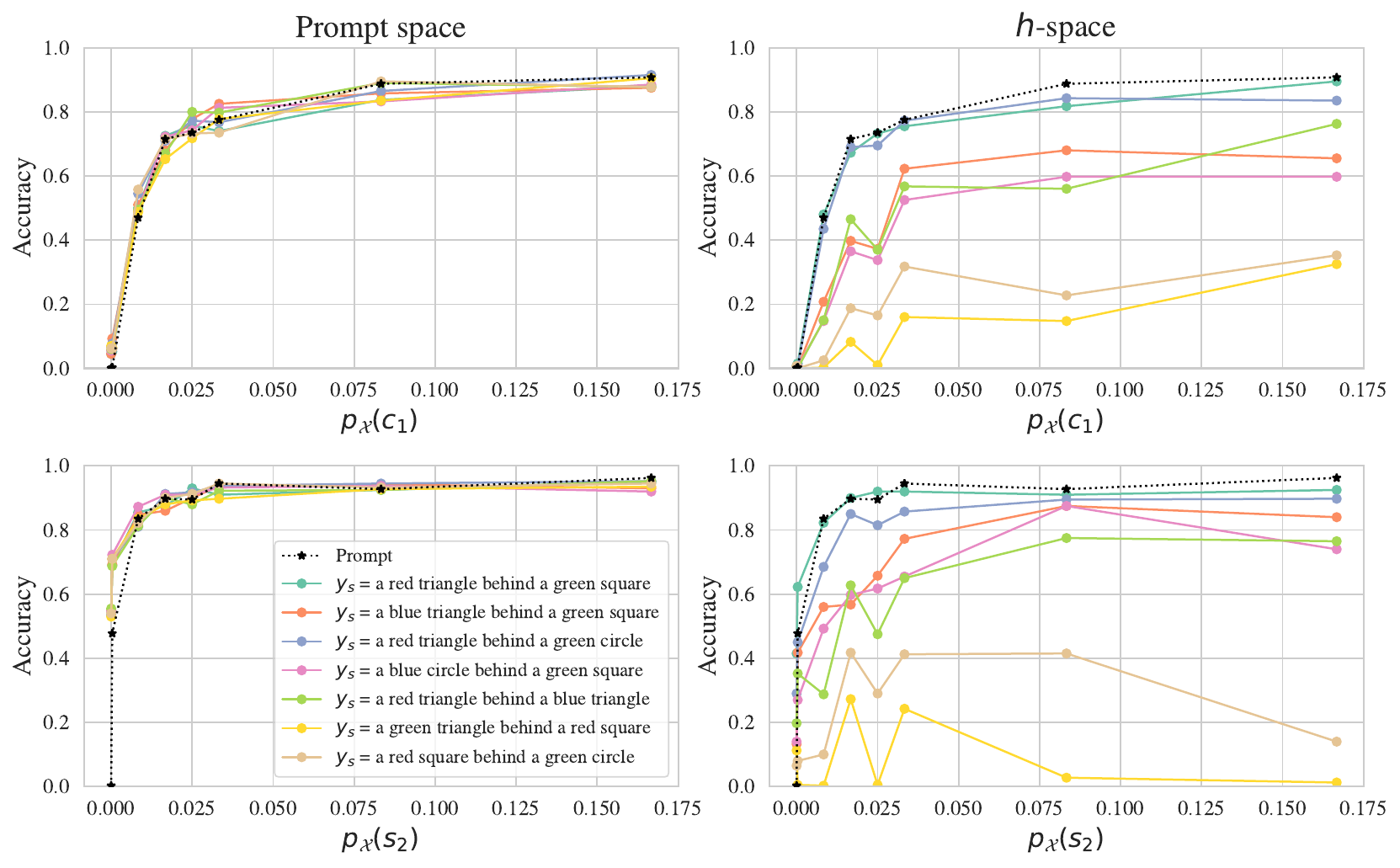}
    \caption{Accuracy of prompting and steering on the prompt space and $h$-space for different starting prompts $y_s$ and varying level of the proportion $p_\mathcal{X}$ of images in the train set containing the concepts $c_1 = red$ (top) and $s_2 = square$ (bottom) across the dataset. The target concept combination for all $y_s$ is $(red, triangle, green, square)$.}
    \label{fig:exp2-comparison_h_p}
\end{figure}

In this section we present additional results for Section \ref{sec:number}. Figure \ref{fig:exp2-comparison_h_p} shows reachability results for different starting prompts $y_s$ when reducing $[c_1 = red]_\mathcal{X}$ and $[s_2 = square]_\mathcal{X}$. As observed in the baseline analysis, steering on the prompt space presents a more stable outcome, while steering on the $h$-space shows greater variability. The highest reachability values on the $h$-space are achieved on the starting prompt $y_s =$ ``a red triangle behind a green square", that mentions the same concepts as those in the target combination. Note that despite the prompt $y_s =$ ``a red triangle behind a green circle" achieving comparable reachability in the case of reduction of $c_1$, this is not consistent in the reduction of the presence of other concepts.

\begin{figure}[htbp]
    \centering
    \includegraphics[width=1\columnwidth]{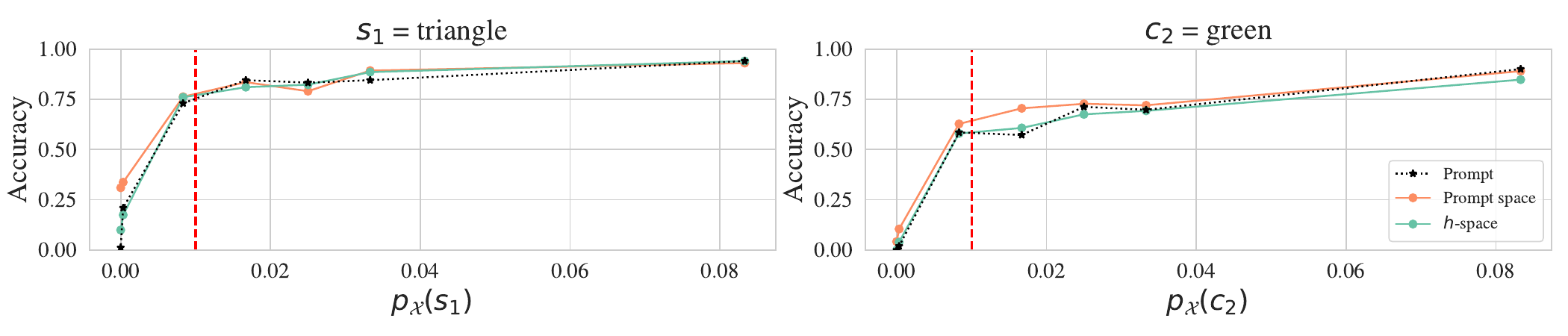}
    \caption{Accuracy of prompting and steering on the prompt space and $h$-space to $(red, triangle, green, square)$ for starting prompt $y_s = $ ``a red triangle behind a green square" and varying the proportion $p_\mathcal{X}$ of images in the training set containing the concepts $s_1 = triangle$ (left) and $c_2 = green$ (right) respectively. The vertical red line marks the approximate threshold 0.01 that determines the approximate shift in reachability.}
    \label{fig:exp2-threshold-add}
\end{figure}

We also present the effect of decreasing the size of the subsets of the dataset $\mathcal{X}$: $[s_1 = triangle]_\mathcal{X}$, $[c_2 = green]_\mathcal{X}$. Figure \ref{fig:exp2-threshold-add} shows a similar sudden decrease in reachability, as observed Figure \ref{fig:exp2-threshold}. %Additional examples from different starting prompts $y_s$ are presented in Figure \ref{fig:exp2-comparison-add}.
Moreover, we highlight the improvement of reachability by steering on the prompt space over prompting below the threshold, most noticeably when reducing the concept $s_1 = triangle$.

%\begin{figure}[htbp]
%    \centering
%    \includegraphics[width=1\columnwidth]{main_plots/exp2/comparison_h_p_add.png}
%    \caption{Accuracy of prompting and steering on the prompt space and $h$-space for different starting prompts $y_s$ and varying level of the presence the concepts $s_1 = triangle$ and $c_2 = green$ across the dataset.}
%    \label{fig:exp2-comparison-add}
%\end{figure}

\subsection{Additional Results for Underspecification} \label{app:label_results}

\begin{figure}[htbp]
    \centering
    \includegraphics[width=0.7\textwidth]{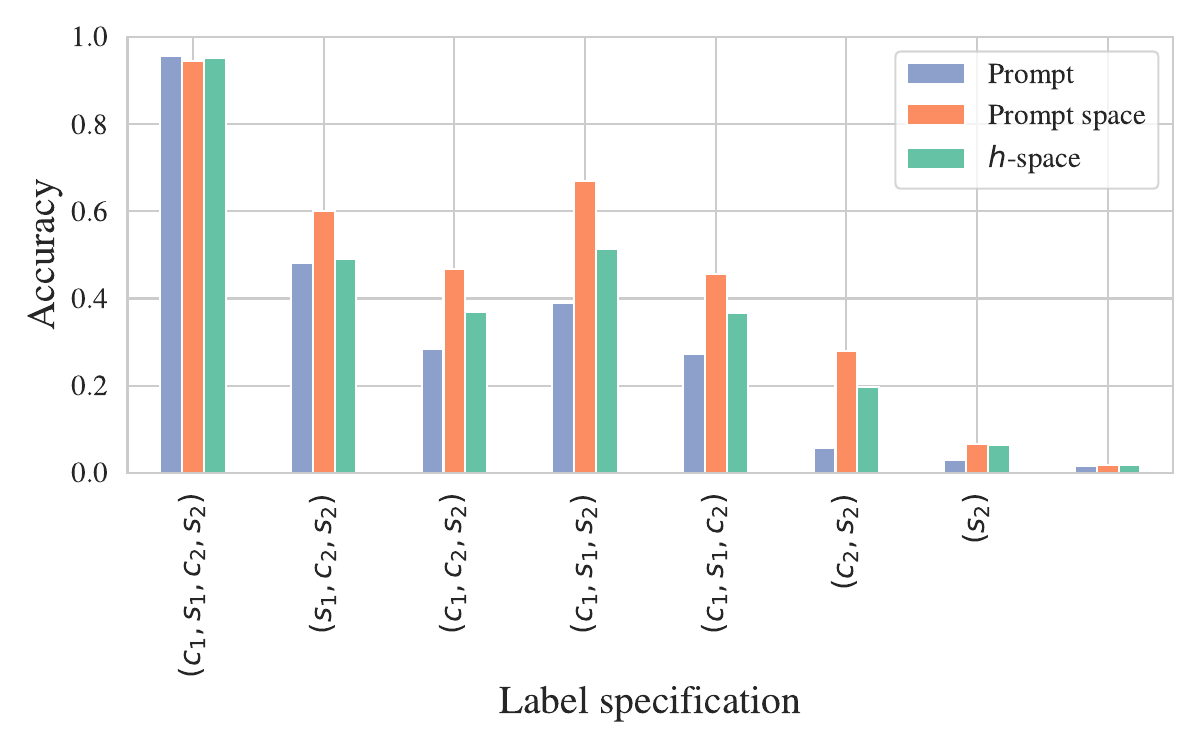}
    \caption{Average reachability across 10 randomly chosen concept combinations $(c_1, s_1, c_2, s_2)$ and varying levels of specification in the captions. Starting prompts $y_s$ used to steer to each of the 10 concept combinations are chosen to fully describe the target concept combinations.}
    \label{fig:exp3-reachability-add}
\end{figure}

In this section we present additional results for analysing the effect of varying the number of label specification on reachability (Section \ref{sec:specification}). We implement steering from the same 10 randomly chosen target concept combinations and trained models as in Figure \ref{fig:exp3-reachability}, but with a starting prompt containing the full semantic information ``a \{$c_1$\} \{$s_1$\} behind a \{$c_2$\} \{$s_2$\}". This approach ensures that the size of the vector $\mathbf{v}_p$ used for steering on the prompt space remains constant for all the models, regardless of the specification level used during training. Figure \ref{fig:exp3-reachability-add} demonstrates that the reachability results show almost no difference to those in Figure \ref{fig:exp3-reachability}, indicating that models are robust to variations of the size of $\mathbf{v}_p$ when steering on the prompt space.

%A visualisation of the classification of the outputs as in Figure \ref{fig:exp3-labels-1} also shows that generated images tend to be unsuccessful in reaching the target combination due to the inability of the model to generate the correct concept for the missing factor. Although the reachability in the prompt space does not show 

%\begin{figure}[htbp]
%    \centering
%    \includegraphics[width=0.95\textwidth]{main_plots/exp3/labels_1_add.png}
%    \caption{Output produced by the model when steering from $y_S = $ ``a red circle behind a green square" towards $(red, circle, green, square)$ when the factors $c_1$, $s_1$, $c_2$ and $s_2$ are removed from the labels in the train set. Each diagram shows the target concept in the top axis, and the alternative concepts the removed factor can take in the remaining directions. The label $X$ represents any other output, including those that do not produce the target combination for the factors seen during training. A model that correctly generates the target concept will produce a high proportion of images on the top axis. If a model only fails to generate the removed concept value, the proportion of images at $X$ will be low.}
%    \label{fig:exp3-labels-1-add}
%\end{figure}

\subsection{Additional Results for Biases} \label{app:tied_concepts}

\begin{figure}[htb]
    \centering
    \includegraphics[width=0.65\columnwidth]{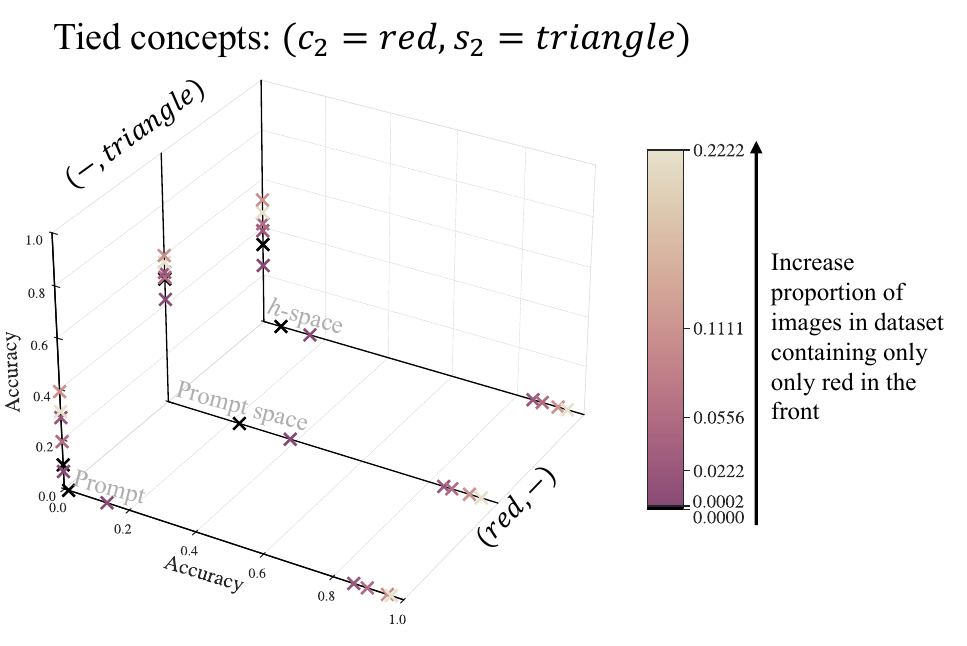}
    \caption{Reachability to concept combinations containing only non-red triangles in the front (vertical axis) and only red non-triangular shapes in the front (horizontal axis), by prompting, steering on the prompt space and steering on the $h$-space. Results are averaged over multiple 6 randomly chosen target concept combinations. When steering, the starting prompt $y_s$ describing those same concepts is used.}
    \label{fig:exp4-start-end-add}
\end{figure}

Figure \ref{fig:exp4-start-end-add} provides an additional example of the scenario studied in Section \ref{sec:tied_concepts}. In particular, we tie the concepts $c_2 = red$ and $s_2 = triangle$, and gradually increase the images in the train set containing non-triangular red shapes. Similar to Figure \ref{fig:exp4-start-end}, we observe a sharp increase in reachability for red shapes as their representation in the dataset grows.

Additionally, although with more variability than Figure \ref{fig:exp4-start-end}, we note a general improvement across all reachability methods on the remaining tied concept (non-red triangles). This suggests that the model learns to disentangle the concept $s_2 = triangle$ as the presence of $c_2 = red$ increases. Moreover, steering consistently outperforms prompting in disentangling the concept $s_2 = triangle$, demonstrating it is a more effective reachability method in such biased scenarios.

\section{Stable Diffusion Implementations}

\subsection{Hyperparameters}

All experiments on real data are conducted on Stable Diffusion v1.5. We use a DDPM scheduler with $T=1000$ inference steps to generate images.

Steering is implemented on the prompt space, with $y_s$ describing the target (e.g., ``an image of a beaker", ``an image of a person walking left"), and a collection $\mathcal{Z}$ of 200 images containing the correct target concepts. The concept vector is initialised at the zero vector and is optimised using Adam for 13 steps. The images required for steering are obtained from openly available datasets such as ImageNet \citep{imagenet} and images sampled from Stable Diffusion and DALLE \citep{dalle2}.

\subsection{Train Dataset} \label{app:train_stable}

Stable Diffusion is primarily trained on subsets of the LAION5B and LAION2B-en datasets \citep{laion}. To estimate the presence of concepts within the dataset, we use unigram and bigram frequencies of images labeled with English captions (LAION2B-en), as provided by \citet{steer_seed}.

The approximate number of captions containing the word \emph{beaker} is 52,000. Despite beakers being easily recognisable and composed of simple visual features, the model struggles to generate them, as shown in Figure \ref{fig:stable_diff}. We hypothesise that this is due to the relatively low number of images containing beakers compared to the overall dataset size.

The bigrams \emph{walking left} and \emph{walking right} appear in only 431 and 585 captions, respectively, whereas the unigram \emph{walking} appears in over 3,450,000 captions—a significantly higher occurrence. Despite this imbalance, the model can occasionally generate images of people facing either left or right, suggesting an inherent understanding of orientation. However, we hypothesise that the lack of explicit directional specification in most captions limits the model’s ability to reliably generate images of people walking in a specific direction.

Finally, the model struggles to generate black-and-white images of rainbows, despite successfully handling black-and-white colour representation in other contexts. We hypothesise that this stems from the scarcity of black-and-white rainbow images in the dataset, making it difficult for the model to disentangle colour information from the broader concept of a rainbow, ultimately revealing a bias in the model’s latent space.

\subsection{Steering on the Prompt Space} \label{app:stable_steer_prompt}

We compare the images generated through sampling using prompting and steering on the prompt space, ensuring both methods use the same initial random seed. This evaluation is conducted on 50 images per studied concept. Below, we present seven examples illustrating the impact of steering on the final output.

\subsubsection{Scarcity of Concepts}

\begin{figure}[htbp]
    \centering
    \includegraphics[width=1\columnwidth]{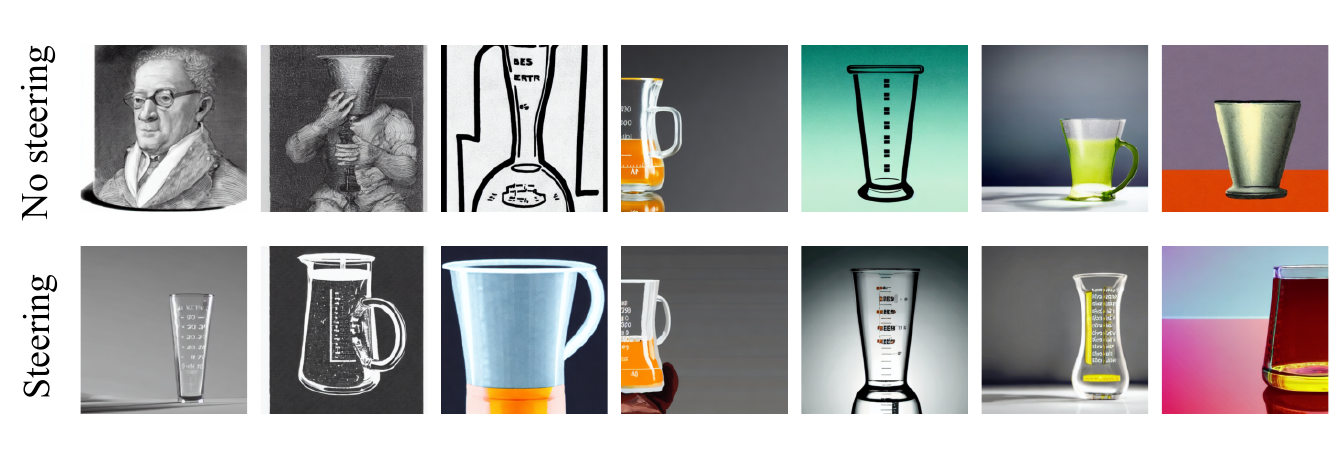}
    \caption{Comparison between sampling on Stable Diffusion with no steering (top row) and with additional steering on the prompt space (bottom row) by using the same random seed. Images are sampled from the prompt ``an image of a beaker". The steering vector is optimised between the same starting prompt $y_s = $ ``an image of a beaker"  and 200 images of beakers.}
    \label{fig:app_beaker}
\end{figure}

Figure \ref{fig:app_beaker} illustrates the effect of steering on image generation towards the target concept: a beaker. We observe that some random seeds initially generate images unrelated to the beaker concept but are successfully guided towards it. In contrast, cases where the model already produces a beaker show minimal modification. Overall, the steered images exhibit attributes commonly associated with beakers, such as transparency, a cylindrical shape, measurement markings, and the presence of liquid.

\subsubsection{Underspecification of Captions}

\begin{figure}[htbp]
    \centering
    \includegraphics[width=1\columnwidth]{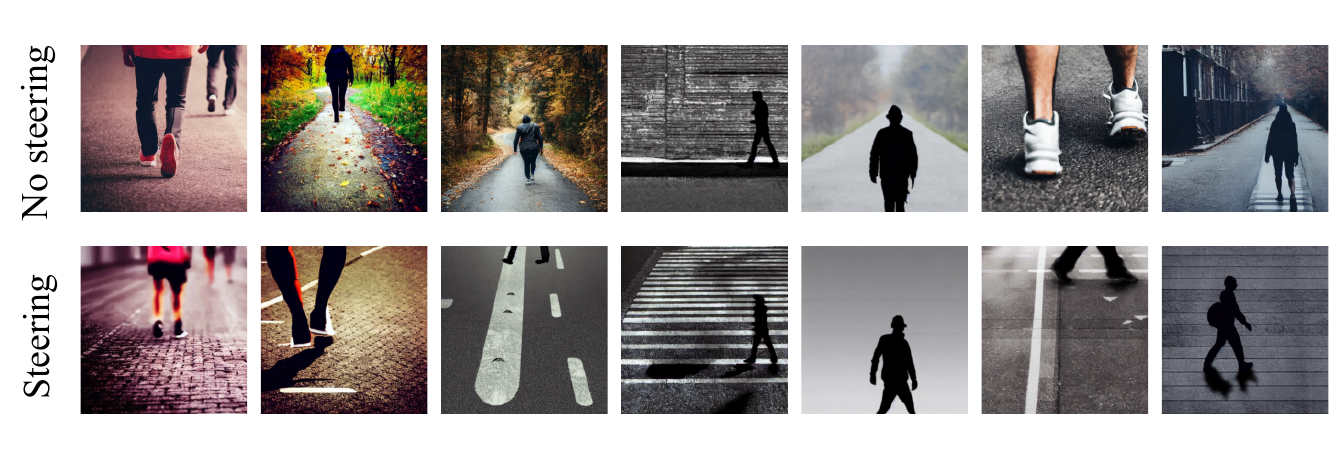}
    \caption{Comparison between sampling on Stable Diffusion with no steering (top row) and with additional steering on the prompt space (bottom row) using the same random seed. Images are sampled from the prompt ``an image of a person walking left". The steering vector is optimised between the starting prompt $y_s = $ ``an image of a person walking left"  and 200 images containing an individual walking towards the left side of the image frame.}
    \label{fig:app_direction}
\end{figure}

Figure \ref{fig:app_direction} illustrates the effect of steering towards generating images of a person walking leftward. Our results indicate that steering in this scenario is limited. In many cases, applying steering on the prompt space does not significantly alter the individual’s position compared to images generated without steering. While we occasionally observe improvements in the walking direction towards the left, we also find instances where the position shifts towards the right—despite no such occurrences in the images used to optimise the steering vector. These findings suggest that reachability is highly constrained when attempting to steer towards concepts that are not explicitly mentioned in the captions. This observation aligns with our discussion in Section \ref{sec:specification}.

\subsubsection{Biases}

\begin{figure}[htbp]
    \centering
    \includegraphics[width=1\columnwidth]{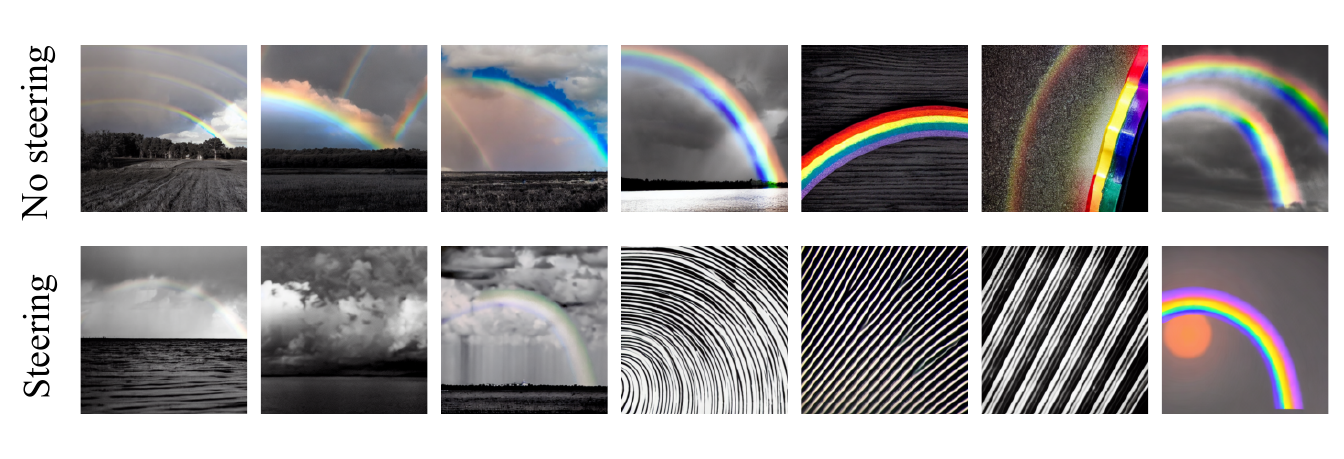}
    \caption{Comparison between sampling on Stable Diffusion with no steering (top row) and with additional steering on the prompt space (bottom row) using the same random seed. Images are sampled from the prompt ``an image of a rainbow in black and white". The steering vector is optimised between the same starting prompt $y_s = $ ``an image of a rainbow in black and white" and 200 greyscale images containing scenes of rainbows.}
    \label{fig:app_rainbow}
\end{figure}

Figure \ref{fig:app_rainbow} illustrates the behaviour of images when steered towards disentangling rainbows from colour. We observe cases where the model reduces the presence of the rainbow, fading the colours from the image, and at times completely removing it. Other images result in an arc-like black-and-white pattern. Interestingly, some steered images resemble black-and-white woodgrain patterns, suggesting a potential bias in the latent space. This may arise because both rainbows and woodgrain share a structure of contrasting light and dark lines, which the model’s latent representations might conflate. Finally, in other instances steering proves ineffective, and the model continues to generate images featuring a coloured arc. 

\section{CelebA} \label{app:celeba}

We conduct a similar analysis to that presented in Section \ref{sec:reachability} using synthetic data, this time using a subset of 16,000 images of faces from the CelebA dataset \citep{celeba}. Leveraging the dataset's existing attribute labels, we assume that the images are generated by underlying factors that take on specific values (concepts), following the structure outlined in Section \ref{sec:concepts}. We note that certain factors and biases present in the dataset may remain unaccounted for, as attributes such as posture or lighting are not labelled in the dataset. Furthermore, the positional relations investigated in our earlier experiments that support positional generalisation are not applicable in this context, which may cause deviations from the behaviours observed in synthetic data. Further challenges may also arise from the use of more entangled and context-dependent words in comparison to the simpler captions in the controlled synthetic setup. Nevertheless, this analysis remains insightful for understanding the model's behaviour in more realistic settings.

\subsection{Hyperparameters}
We study the attributes gender ($g \in \{ \text{man}, \text{woman} \}$) and hat ($\hat{h} \in \{ \text{wearing a hat}, \text{without a hat} \}$). Captions are composed in the form ``a $\{g\}$ $\{\hat{h}\}$". For example, ``a woman wearing a hat", or ``a man without a hat". The balanced dataset is comprised of 4,000 images of each of the four possible concept combinations. The images are transformed using a random horizontal flip ($p=0.5$) and colour jitter with brightness 0.1, contrast 0.1 and saturation 0.1. As in our previous experiments, when varying the concepts in the train set we approximately preserve the total dataset size. The model architecture is the same as described in Appendix \ref{app:hyperparameters}. We train our models for 400 epochs using the same optimiser and learning rate as before.

The concept vector for steering on the prompt space and $h$-space is obtained using Adam with learning rate 0.02 for 11 steps, and 100 images containing the target concept combinations. To evaluate reachability, we train 4 diffusion models with different random seeds, use them to generate 100 images of the target concept combinations and report the mean results. To help in the evaluations of the generated images, we train two CNNs consisting of three convolutional layers and two linear layers to classify (i) the gender of the person in the image and (ii) if they are wearing a hat or not. Our classifiers obtain an accuracy of $94.6\%$ (gender) and $97.2\%$ (hat) on a held-out validation set consisting of 2,400 images.

\subsection{Baseline}

\begin{figure}[htbp]
    \centering
    \begin{subfigure}{0.36\textwidth}
        \includegraphics[width=\linewidth]{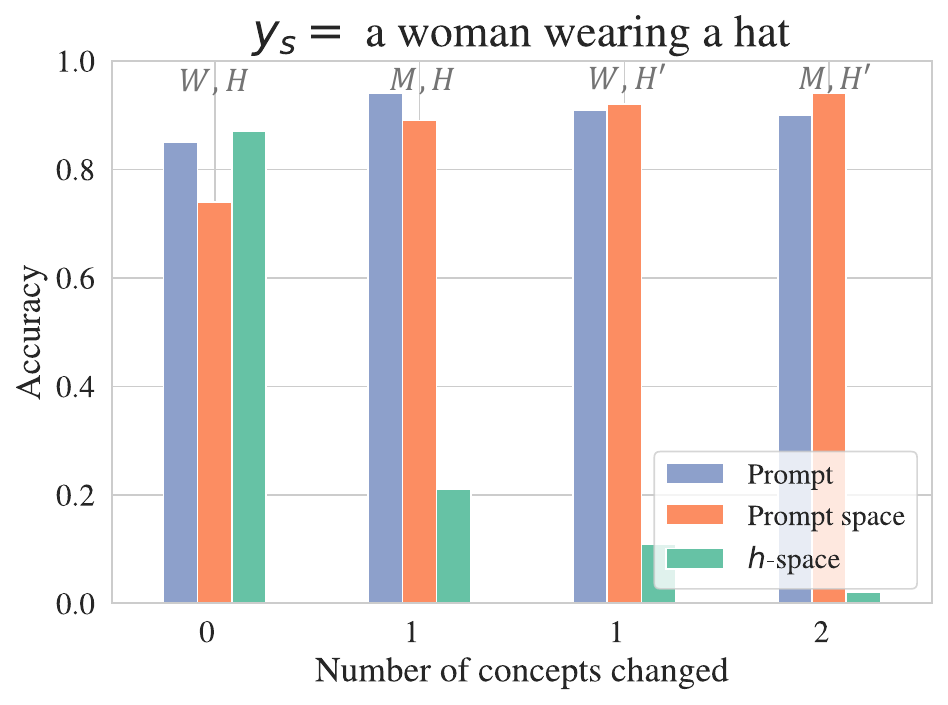}
        \caption{} % No individual caption
        \label{fig:celeba_exp1}
    \end{subfigure}
    \hfill
    \begin{subfigure}{0.57\textwidth}
        \includegraphics[width=\linewidth]{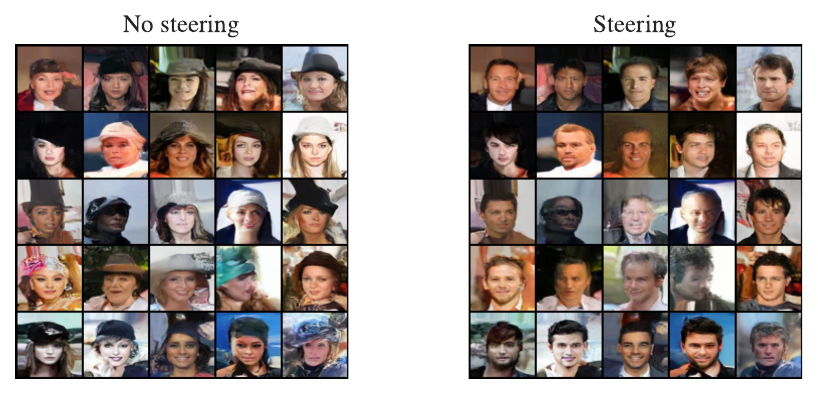}
        \caption{} % No individual caption
        \label{fig:celeba_exp1_examples}
    \end{subfigure}
    \caption{a) Reachability to different concept combinations when prompting and when steering from the starting prompt $y_s = $``a woman wearing a hat". Target combinations are shown at the top of each bar, and are organised according to the number of concepts that differ from the concepts of $y_s$. The notation $M$/$W$ refers to the gender (man/woman) and $H$/$H'$ to wearing a hat/without a hat. b) Example of images generated from the prompt ``a woman wearing a hat" (no steering) and the images obtained by additionally steering on the prompt space to the concept combination $(\text{man}, \text{without a hat})$.}
\end{figure}

Fixing a starting prompt $y_s$, we apply steering towards the different target concept combinations. For comparison, we also evaluate the reachability achieved through direct prompting of the target combinations. The results are presented in Figure \ref{fig:celeba_exp1}. Additionally, Figure \ref{fig:celeba_exp1_examples} illustrates a comparison of images generated by the model under no steering (prompting $y_s = $``a woman wearing a hat") and with additional steering towards the concept combination $(\text{man}, \text{without a hat})$.

%\begin{figure}[htbp]
%    \centering
%    \includegraphics[width=0.5\columnwidth]{main_plots/celeba/exp1.png}
%    \caption{Reachability to different concept combinations when prompting, as well as steering from the starting prompt $y_s = $``a woman wearing a hat". Target combinations are shown at the top of each bar, and are organised according to the number of concepts that differ from the concepts of $y_s$. The notation $M$/$W$ refers to the gender (man/woman) and $H$/$H'$ to wearing a hat/without a hat.}
%    \label{fig:celeba_exp1}
%\end{figure}

\paragraph{Concepts remain reachable when steering from diverse starting prompts}
Figure \ref{fig:celeba_exp1} demonstrates that the behaviour observed under balanced conditions aligns with the patterns identified in the synthetic setting. Specifically, while concepts generally remain reachable through prompt-space steering, reachability in the $h$-space declines as the number of altered concepts increases. Overall, steering enables effective access to target concepts, achieving a level of reachability comparable to that of directly prompting the desired concepts. Notably, we observe a systematic decrease in reachability for the concept combination $(\text{woman}, \text{wearing a hat})$ relative to the other combinations, which may reflect latent biases in the training data distribution.

\subsection{Scarcity of Concepts}

\begin{figure}[htbp]
    \centering
    \includegraphics[width=0.6\columnwidth]{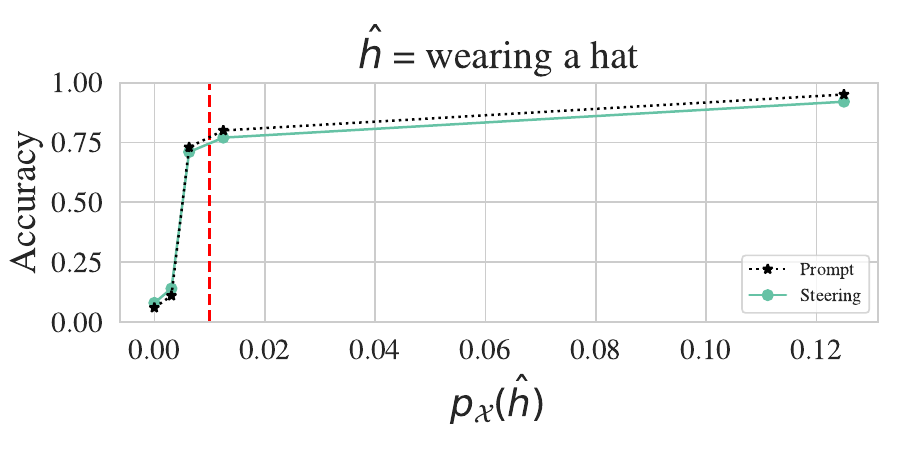}
    \caption{Accuracy of prompting and steering from the starting prompt $y_s = $ ``a man wearing a hat" for a varying proportion $p_\mathcal{X}$ of images across the dataset containing the concept $\hat{h}=\text{wearing a hat}$. The steering curve shows the results for $\max\text{(prompt space}, h\text{-space})$. %Images containing the target combination were generated through a diffusion model trained on the balanced dataset.
    The dotted red line marks the approximate threshold 0.01 of the shift in reachability.}
    \label{fig:celeba_exp2}
\end{figure}

We decrease the presence of the concept $\hat{h} = \text{wearing a hat}$, and consider reachability to the concept combination $(\text{man}, \text{wearing a hat})$. Results are presented in Figure \ref{fig:celeba_exp2}.

\paragraph{Reachability drops sharply past a critical threshold} Figure \ref{fig:celeba_exp2} reveals a threshold pattern consistent with that observed in Figure \ref{fig:exp2-threshold}. Specifically, we observe a sharp decline in reachability once the proportion of training images containing a given concept falls below a certain threshold. As in the synthetic setting, the model appears able to learn and generalize a concept even from a small number of examples, provided this minimal threshold is met. This highlights the importance of ensuring minimal representation of key concepts in training data to ensure reachability.

\subsection{Underspecification of Concepts}

We vary the level of specification of captions of the training dataset, and in particular reduce the captions of the form ``a $\{g\}$ $\{\hat{h}\}$" to the following:
\begin{enumerate}
    \item Remove $g$: replace the caption with ``a person $\{\hat{h}\}$"
    \item Remove $\hat{h}$: replace the caption with ``a $\{g\}$"
    \item Remove $g$ and $\hat{h}$: replace the caption with the empty string, ``"
\end{enumerate}

\begin{figure}[htbp]
    \centering
    \begin{subfigure}{0.45\textwidth}
        \includegraphics[width=0.9\textwidth]{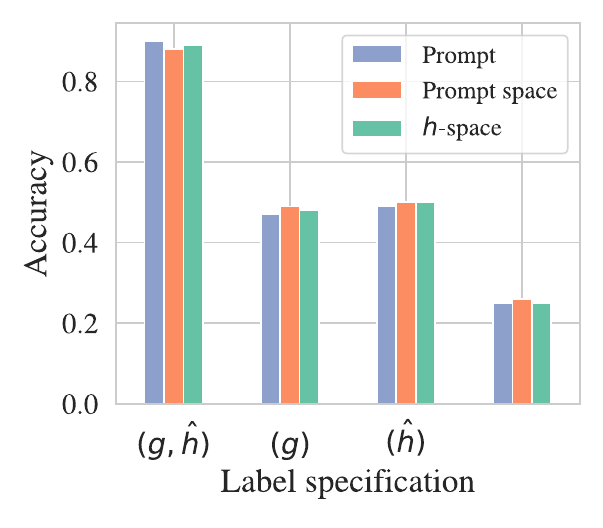}
        \caption{}
        \label{fig:celeba_exp3}
    \end{subfigure}
    \hfill
    \begin{subfigure}{0.5\textwidth}
        \includegraphics[width=\linewidth]{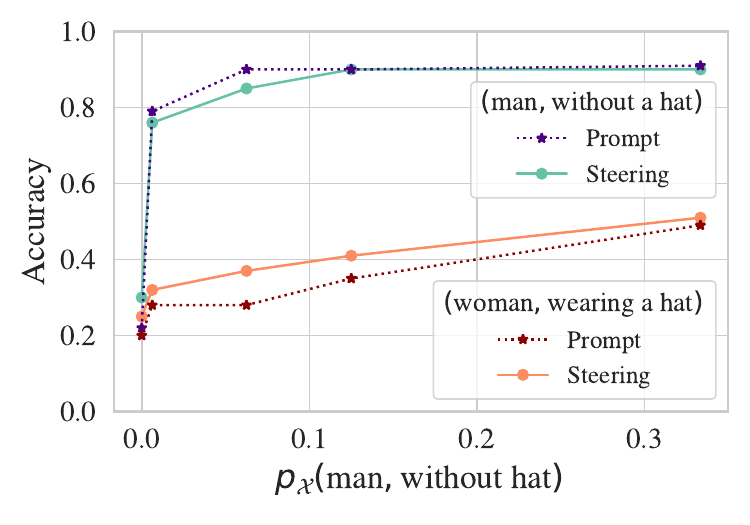}
        \caption{}
        \label{fig:celeba_exp4}
    \end{subfigure}
    \caption{a) Average accuracy of prompting and steering to the different concept combinations for different levels of concept specification. b) Accuracy of prompting and steering to the concept combinations $(\text{man}, \text{without a hat})$ and $(\text{woman}, \text{wearing a hat})$ under biased conditions. Starting prompts $y_s$ describe the target concept combinations, while varying the presence of the concept combination $(\text{man}, \text{without a hat})$. The steering curve shows the results for $\max\text{(prompt space}, h\text{-space})$.}
\end{figure}

We steer from the starting prompt describing only seen concepts. For example, when removing $\hat{h}$ from the captions, to steer to $(\text{woman}, \text{wearing a hat})$ we use $y_s=$ ``a woman". Additionally, we compare the accuracy of prompting using the complete description of the target concepts. Results are presented in Figure \ref{fig:celeba_exp3}.

%\begin{figure}[htbp]
%    \centering
%    \includegraphics[width=0.5\columnwidth]{main_plots/celeba/exp3.png}
%    \caption{Accuracy of prompting and steering from the starting prompt $y_s = $ ``a man wearing a hat" for a varying proportion $p_\mathcal{X}$ of images across the dataset containing the concept $\hat{h}=\text{wearing a hat}$. The steering curve shows the results for $\max\text{(prompt space}, h\text{-space})$. %Images containing the target combination were generated through a diffusion model trained on the balanced dataset. The dotted red line marks the approximate threshold 0.01 of the shift in reachability.}
%    \label{fig:celeba_exp3}
%\end{figure}

\paragraph{Underspecification hinders reachability} Similar to Section \ref{sec:specification}, we observe a rapid decrease in the reachability of concepts. The reachability is observed to be close to the value of sampling the unspecified concepts randomly: when one factor is unspecified, reachability is approximately 50\%, and when both are unspecified, reachability is approximately 25\%. %We also observe that when concepts are underspecified, there is a slight improvement in reachability when steering.

\subsection{Biases}

We tie the concepts ``man" and ``wearing a hat" (which also causes the concepts ``woman" and ``without a hat" to be tied), and gradually increase the presence of images containing the concept combination $(\text{man}, \text{without a hat})$, thus reducing the bias. We evaluate reachability to $(\text{man}, \text{without a hat})$ and $(\text{woman}, \text{wearing a hat})$. Results are presented in Figure \ref{fig:celeba_exp4}.

%\begin{figure}[htbp]
%    \centering
%    \includegraphics[width=0.65\columnwidth]{main_plots/celeba/exp4.png}
%    \caption{Accuracy of prompting and steering to the concept combinations $(\text{man}, \text{without a hat})$ and $(\text{woman}, \text{wearing a hat})$. Starting prompts $y_s$ describe the target concept combinations. The steering curve shows the results for $\max\text{(prompt space}, h\text{-space})$.}
%    \label{fig:celeba_exp4}
%\end{figure}

\paragraph{Increasing the presence of an individual concept increases separate reachability to both concepts} Figure \ref{fig:celeba_exp4} shows a clear increase in reachability to either concept combination as the bias in the dataset is decreased. Most noticeably, we observe a sudden increase in reachability to $(\text{man}, \text{without a hat})$, similar to the threshold patterns previously observed. Moreover, the reachability to $(\text{woman}, \text{wearing a hat})$ also gradually increases, suggesting that the models become more capable of disentangling concepts. We further note that this concept combination is consistently more reachable through steering.

\end{document}